\documentclass[12pt]{article}
\usepackage{thesis}
\usepackage{geometry}
\usepackage{soul,xcolor}
\setstcolor{red}

\newcommand{\remove}[1]{}

\usepackage{algorithm}
\makeatletter
\renewcommand{\fnum@algorithm}{\fname@algorithm}
\makeatother
\usepackage{amsmath}
\usepackage{amsthm}
\usepackage{amssymb}
\usepackage{amsfonts}
\usepackage[inline]{enumitem}
\usepackage[flushleft]{threeparttable}
\usepackage{array,multirow}

\usepackage{tikz}
\usetikzlibrary{shapes.geometric, arrows}
\usepackage{caption}

\usepackage{changepage}
\usepackage{setspace}
\usepackage[en-US]{datetime2}

\usepackage{hyperref}
\usepackage[noadjust]{cite}

\newcommand\fnote[1]{\captionsetup{font=small}\caption*{#1}}

\newtheorem{theorem}		{Theorem} [section]
\newtheorem{lemma}			[theorem]	{Lemma}
\newtheorem{definition}		[theorem]	{Definition}
\newtheorem{corollary}		[theorem]	{Corollary}
\newtheorem{Claim}			[theorem]	{Claim}
\newtheorem{question}		[theorem]	{Question}

\DeclareMathSymbol{\N}{\mathbin}{AMSb}{"4E}
\DeclareMathSymbol{\Z}{\mathbin}{AMSb}{"5A}
\DeclareMathSymbol{\R}{\mathbin}{AMSb}{"52}
\DeclareMathSymbol{\Q}{\mathbin}{AMSb}{"51}
\DeclareMathSymbol{\erert}{\mathbin}{AMSb}{"50}
\DeclareMathSymbol{\I}{\mathbin}{AMSb}{"49}
\DeclareMathSymbol{\C}{\mathbin}{AMSb}{"43}

\newcommand{\cost}{{\rm cost}}
\def\opt{\mathop{\rm{OPT}}\nolimits}
\def\plainopt{\mathop{\rm{OPT}}\nolimits}

\def\argmin{\mbox{\rm argmin}}
\def\view{\mbox{\rm View}}
\newcommand{\poly}{\mathop{\rm poly}}

\makeatletter
\newlength{\fboxhsep}
\newlength{\fboxvsep}
\newlength{\fboxtoprule}
\newlength{\fboxbottomrule}
\newlength{\fboxleftrule}
\newlength{\fboxrightrule}
\def\@frameb@xother#1{%
  \@tempdima\fboxtoprule
  \advance\@tempdima\fboxvsep
  \advance\@tempdima\dp\@tempboxa
  \hbox{%
    \lower\@tempdima\hbox{%
      \vbox{%
        \hrule\@height\fboxtoprule
        \hbox{%
          \vrule\@width\fboxleftrule
          #1%
          \vbox{%
            \vskip\fboxvsep
            \box\@tempboxa
            \vskip\fboxvsep}%
          #1%
          \vrule\@width\fboxrightrule}%
        \hrule\@height\fboxbottomrule}%
    }%
  }%
}
\long\def\fboxother#1{%
  \leavevmode
  \setbox\@tempboxa\hbox{%
    \color@begingroup
    \kern\fboxhsep{#1}\kern\fboxhsep
    \color@endgroup}%
  \@frameb@xother\relax}

\makeatother

\newcommand{\Lap}{\operatorname{\rm Lap}}

\newcommand{\AAA}{\mathcal A}
\newcommand{\BBB}{\mathcal B}

\newcommand{\CCC}{\mathcal C}
\newcommand{\DDD}{\mathcal D}

\newcommand{\NNN}{\mathcal N}

\newcommand{\XXX}{\mathcal X}
\newcommand{\eps}{\varepsilon}

\newcommand{\wdist}{d_{\sf W}}
\newcommand{\polylog}{\mathop{\rm polylog}}
\newcommand{\medopt}{{\rm OPT}^{1}}
\newcommand{\medcost}{{\rm cost}^{1}}
\newcommand{\Med}{{\rm med}}
\tikzstyle{stability_assumption} = [rectangle, rounded corners, minimum width=3cm, minimum height=0.8cm,text centered, draw=black, fill=gray!10]
\tikzstyle{implication}      = [ultra thick,->,>=stealth]
\tikzstyle{implication_weak} = [ultra thick,->,>=stealth]

\begin{document}

%%%%%%%%%%%%%%%%%%%%%%%%%%%%%%%%%%%%%%%%%%%%%
%% Cover code - BGU official
%%%%%%%%%%%%%%%%%%%%%%%%%%%%%%%%%%%%%%%%%%%%%
\pagenumbering{roman}
\pagestyle{empty}
	
\DTMlangsetup{showdayofmonth=false}
\thesistitle
	{Differentially Private Algorithms for\\
Clustering with Stability Assumptions}
	{Moshe Shechner}
%	\date{\normalsize\today}
	\date{January 2021}
	
\cleardoublepage %skip the next page

\begin{center}
\large
Ben-Gurion University of the Negev \\
Faculty of Natural Sciences \\
Department of Computer Science \\

\vspace{\fill}
\textbf{\LARGE
Differentially Private Algorithms for\\
Clustering with Stability Assumptions
}

\large
\vspace{\fill}
Thesis submitted in partial fulfillment of the requirements \\
for the M.Sc. degree in the Faculty of Natural Sciences \\

by \\
Moshe Shechner\\

\vspace{\fill}  
Under the supervision of\\ Dr. Uri Stemmer \\
\end{center}
\vspace{2.0cm}
\begin{flushleft}

Author signature\underline{\qquad \qquad \qquad \qquad \qquad \qquad}
\hfill 
Date\underline{\qquad \qquad \qquad \qquad}\\
Advisor approval\underline{\qquad \qquad \qquad \qquad \qquad \qquad}
\hfill 
Date\underline{\qquad \qquad \qquad \qquad}\\
Head of teaching committee approval\underline{\qquad \qquad \qquad \qquad}
\hfill 
Date\underline{\qquad \qquad \qquad \qquad}
\end{flushleft}
\setlength{\baselineskip}{18pt}	
\cleardoublepage

\setlength{\baselineskip}{18pt}	
\pagestyle{plain}
\clearpage
%%%%%%%%%%%%%%%%%%%%%%%%%%%%%%%%%%%%%%%%%%%%%
%% end of cover code
%%%%%%%%%%%%%%%%%%%%%%%%%%%%%%%%%%%%%%%%%%%%%

%%%%%%%%%%%%%%%%%%%%%%%%%%%%%%%%%%%%%%%%%%%%
% Abstract
%%%%%%%%%%%%%%%%%%%%%%%%%%%%%%%%%%%%%%%%%%%%
\addcontentsline{toc}{section}{Abstract}
\begin{center}
\textbf{\huge Abstract}\label{sec:Abstract}
\end{center}
\vspace{1.0cm} 

We study the problem of differentially private clustering under input-stability assumptions. Despite the ever-growing volume of works on differential privacy in general and differentially private clustering in particular, only three works~(Nissim et al.\ 2007, Wang et al.\ 2015, Huang et al.\ 2018) looked at the problem of privately clustering ``nice'' $k$-means instances, all three relying on the sample-and-aggregate framework and all three measuring utility in terms of Wasserstein distance between the true cluster centers and the centers returned by the private algorithm. In this work we improve upon this line of works on multiple axes. We present a far simpler algorithm for clustering stable inputs (not relying on the sample-and-aggregate framework), and analyze its utility in both the Wasserstein distance and the $k$-means cost. Moreover, our algorithm has straight-forward analogues for ``nice'' $k$-median instances and for the local-model of differential privacy.
\clearpage

%%%%%%%%%%%%%%%%%%%%%%%%%%%%%%%%%%%%%%%%%%%%
% Acknowledgment
%%%%%%%%%%%%%%%%%%%%%%%%%%%%%%%%%%%%%%%%%%%%
\addcontentsline{toc}{section}{Acknowledgments}

\begin{center}
\textbf{\huge Acknowledgments}
\end{center}

\vspace{1.0cm}
This work concludes a first and enjoyable step in research that took place in the last two years. 
First and foremost, I would like to thank my advisor, Dr.\ Uri Stemmer, that apart of his professional guidance, has also showed me the joy of intensive research work. I believe that the experience of research in that period of time was enjoyable largely thanks to his facilitation. For that I feel fortunate. I also thank Dr.\ Or Sheffet for the opportunity to work with him. Lastly, I thank my friends and my dear family for the support.
\clearpage

%%%%%%%%%%%%%%%%%%%%
% Table Of Contents
%%%%%%%%%%%%%%%%%%%%
\setlength{\baselineskip}{15pt}
\tableofcontents
\clearpage

%%%%%%%%%%%%%%%%%%%%
%% Chapters 
%%%%%%%%%%%%%%%%%%%%
\onehalfspace
\pagenumbering{arabic}
\pagestyle{plain}

%%%%%%%%%%%%%%%
% Introduction
%%%%%%%%%%%%%%%
\section{Introduction}
In recent years differential privacy~\cite{DBLP:conf/tcc/DworkMNS06} has been established as the de-facto gold standard of privacy preserving data analysis. The notion of differential privacy guarantees that any single datum has a limited effect on the outcome of the algorithm, and so it is often presented as a formal notion of robustness. Indeed, it is commonly believed that objectives which are sensitive to the change of a single datapoint are hard to approximate in a differentially private manner. One such notorious example is the median, which may shift drastically by a single datapoint.
And yet, the median is easy to approximate on ``nice'' or ``stable'' instances (e.g., instances in which there are many datapoints in the vicinity of the median). In fact, the median problem was the first to be studied in the context of the interplay between input-stability notions and the stability enforced by differential privacy.~\cite{NRS07}

In this work we aim to utilize input stability assumptions in order to design differentially private clustering algorithms with improved utility guarantees. We focus on the task of {\em $k$-means clustering.} 
The construction of differentially private $k$-means clustering algorithms has attracted a lot of attention over the last 14 years~\cite{BDMN05,NRS07,FFKN09,
McSherry09,GuptaLMRT10,Mohan2012,Wang2015,NockCBN16,
Su2016,NSV16,DannyPrivatekMeans,Balcan17a,
NS18_1Cluster,HuangL18,KaplanSt18,Stemmer20}. In particular, three of these works~--- the work of Nissim et al.~\cite{NRS07} and two followup papers~\cite{Wang2015,HuangL18}~--- have constructed private $k$-means algorithms for stable instances. 
%Recently, Huang and Liu~\cite{HuangL18} gave a revised version of the sample-and-aggregate base approach of Nissim et al, relying on the 1-cluster algorithm~\cite{NSV16, NS18_1Cluster}. 
While several interesting concepts arise from these three works, their algorithms~--- and more importantly, their analysis~--- can be tighten up, simplified, and at the same time be applied in a broader setting. Our work does precisely this: we simplify the existing constructions for private clustering on stable instances, while improving upon their analysis and relating it to several different notions of input stability, applicable for both the $k$-means and the $k$-median objectives. Moreover, we give the first local-differentially private algorithm for clustering stable instances.

\subsection{Our Setting}
Before formally presenting our results, we describe our setting more precisely. Consider an input database $X$ containing $n$ points in $\R^d$. In $k$-means clustering, the goal is to identify a set $C$ of $k$ {\em centers} in $\R^d$, approximately minimizing the sum of squared distances from each input point to its nearest center, a quantity referred to as the {\em cost} of the centers. It is sometimes more suitable to minimize the sum of distances to the centers (instead of squared distances), in which case the problem is called {\em $k$-median clustering}. Formally, for a set of points $X\in(\R^d)^n$, a set of centers $C\subseteq \R^d$, and a parameter $p$, define
$$\cost^p_X(C)=\sum_{x\in X}\min_{c\in C}\|x-c\|^p.$$
In $k$-means clustering we aim to find a set of $k$ centers $C$ minimizing $\cost^2_X(C)$, and in $k$-median clustering we aim to minimize $\cost^1_X(C)$. We use $\opt^2_k(X)$ and $\medopt_k(X)$ to denote lowest possible $k$-means and $k$-median cost of $X$. When the cost objective is clear from the context we drop the superscript and simply write $\cost_X(C)$ and $\opt_k(X)$.

As minimizing the $k$-means and $k$-median objectives is NP-hard~\cite{aloise2009np,dasgupta2009random,drineas2004clustering,DBLP:journals/tcs/MahajanNV12,megiddo1984complexity}, the literature has focused on approximation algorithms, with the current (non-private) state-of-the-art constructions achieving multiplicative error of 6.357 for $k$-means and 2.633 for $k$-median~\cite{ahmadian2019better}. That is, the algorithm of~\cite{ahmadian2019better} for $k$-means identifies a set of $k$ centers whose cost is no more than $6.357\cdot\opt^2_k(X)$.
Furthermore, for {\em stable instances}, one can obtain significantly improved guarantees, with error arbitrarily close to 1 (non-privately)~\cite{DBLP:journals/jacm/OstrovskyRSS12,BalcanBG09, AwasthiBS10, BiluL10, KumarK10,AwasthiBS12}. In this work we focus on the input stability notion of Ostrovsky et al.~\cite{DBLP:journals/jacm/OstrovskyRSS12}, defined as follows.

\begin{definition}[Ostrovsky et al.~\cite{DBLP:journals/jacm/OstrovskyRSS12}]
A clustering instance $X$ is {\em $\phi$-well-separated} (or simply $\phi$-separated) for 
$k$-clustering objective with parameter $p$ if the following holds:
$$ \frac{\plainopt^p_k(X)}{\plainopt^p_{k-1}(X)} \leq \phi^p.$$ 
That is, the optimal clustering cost with $k$ centers is significantly lower then the optimal cost for $k-1$ centers.
\end{definition}

In our context, every input point $x\in X$ is assumed to be the (private) information of one individual (such as a location or a text file), and we would like to identify a set of centers $C$ with low cost while at the same time providing differential privacy for the points in $X$. This means that we are interested in {\em randomized} clustering algorithms that guarantee that their outcome distribution (i.e., the distribution on the returned centers) is insensitive to any arbitrary modification of a single datapoint. Formally,

%The definition of differential privacy is,
\begin{definition}[Differentially private algorithm \cite{DBLP:conf/tcc/DworkMNS06}]\label{def:dp_intro}
Let $\XXX$ be a domain of record type. A randomized algorithm $\AAA : \N^{\XXX} \rightarrow Y$ is $(\eps,\delta)$ {\em differentially private} if for every two databases $X,X'\in \N^{\XXX}$ that differ in one row, and every set $T\subseteq Y$, we have 
$$\Pr[\AAA(X)\in T]\leq e^{\eps}\cdot \Pr[\AAA(X')\in T]+\delta.$$
\end{definition}

Unlike in the non-private literature, it is known that every {\em private} algorithm for approximating the $k$-means must have an {\em additive} error (even computationally unbounded algorithms), which scales with the diameter of the input space. Hence, a standard assumption for private $k$-means is that the input points come from the $d$-dimensional ball of radius $\Lambda$ around the origin $\BBB(0,\Lambda)$. This is the setting we consider in this work, where we fix $\Lambda=1$ for the introduction.
%(even computationally unbounded algorithms, see introduction of~\cite{KaplanSt18})
As private $k$-means algorithms have both multiplicative and additive errors, different guarantees can easily be incomparable. 
Typically (though not always), one aims to minimize the multiplicative error while keeping the additive error at most polylogarithmic in the size of the database (note that an additive error of size $|X|$ is meaningless). The current state-of-the-art construction for private $k$-means by Kaplan and Stemmer~\cite{KaplanSt18} obtained a $O(1)$ multiplicative error and $\poly(\log(n),k,d)$ additive error.

Given the success of (non-private) stability-based clustering algorithms, it is not surprising that such stability assumptions were also utilized in the privacy literature, specifically by~\cite{NRS07,Wang2015,HuangL18}. However, the error measure pursued in these three works is different. Instead of aiming to find $k$ centers with low $k$-means cost, these three works aim to find centers that are close to the optimal centers in terms of the {\em Wasserstein distance}, defined as follows.

\begin{definition}[Wasserstein distance \cite{vaserstein1969markov}]
Let $C=(c_1,\dots,c_k)\in(\R^d)^k$ and $\hat{C}=(\hat{c}_1,\dots,\hat{c}_k)\in(\R^d)^k$ be two sets of centers. The {\em Wasserstein distance} between $C$ and $\hat{C}$ is the $L^{dk}_2$ distance under the best possible permutation $\pi$ of the centers in each set. Denote it by $d_W(C,\hat{C})$.
\end{definition}

Nissim et al.~\cite{NRS07} presented a private algorithm that, for a $\phi$-well separated instance, computes $k$ centers of Wasserstein distance at most $O(\eps^{-1}d\sqrt{k}\cdot\phi^2)$ from the optimal $k$-means centers. Wang et al.~\cite{Wang2015} extended the results of Nissim et al.\ to {\em subspace clustering}\footnote{In subspace clustering we aim to group the data points into clusters so that data points in a single cluster lie approximately on a low-dimensional linear subspace.} with similar error bounds. Finally, Huang and Liu~\cite{HuangL18}, presented a clever algorithm that reduced the error down to $O(\phi^2)$ -- a significant improvement over the previous error bounds of \cite{NRS07} and \cite{Wang2015}. In addition, \cite{HuangL18} showed that their error bound is tight, and that Wasserstein distance of $O(\phi^2)$ is the best possible under differential privacy (for $\phi$-separated instances).
This naturally raises the following question, which is the starting point of our research.
\newpage
\begin{question}
Can input stability assumptions be utilized in order to construct {\em differentially private} clustering algorithms that guarantee low error in terms of the {\em $k$-means cost}  (rather than Wasserstein distance)?
\end{question}

We comment that even though a set of centers $\hat{C}$ might be close to the optimal centers $C$ in terms of the Wasserstein distance, say $\wdist(C,\hat{C})=\gamma$, the $k$-means cost of $\cost_{X}(\hat{C})$ might be as big as $\opt_k(X)+|X|\cdot\gamma^2$. That is, the additive error obtained by translating a bound on the Wasserstein distance to a bound on the $k$-means cost scales with $|X|$. In this work we are aiming for an additive error of at most $\polylog|X|$, which means that approximation guarantees w.r.t.\ the Wasserstein distance do not imply (in general) satisfactory approximation guarantees w.r.t.\ the $k$-means cost.

\subsection{Our Contribution and Organization}
 First, we establish equivalence between several notions of input-stability for clustering problems. This result is given in the preliminaries, Section~\ref{sec:preliminaries}, and should come as no surprise considering all of these notions (and others) yield a PTAS\footnote{PTAS - Polynomial Time Approximation Scheme, is an approximation algorithm that admit the following definition: for any fixed $\alpha > 0$ it finds a solution with a value within a factor of $(1+\alpha)$ from the value of the optimal solution, in polynomial time ($\alpha$ is given as part of the input). We note that typically a PTAS has a runtime of $O(n^{1/\alpha})$ Therefore the power of $n$ may be quite large for small approximation constant.} for the clustering problem (non-privately) \cite{DBLP:journals/jacm/OstrovskyRSS12, AwasthiBS10}. Second, we present our~-- absurdly simple~-- private algorithm for clustering well-separated instances in Section~\ref{sec:main}, which can be summarized as follows: run an arbitrary (private) $k$-means approximation algorithm and then take a Lloyd-step (averaging only the points with clear preference for one center over all others). We give a short proof arguing that the result of applying an algorithm with a worst-case guarantee of $v$-approximation\footnote{With an additive error, as analyzed in Theorem~\ref{thm:main}.} to the $k$-means objective on a $\phi$-well separable instance is (effectively) a $(1+O(\phi^2))$-approximation, provided $v$ is small in comparison to $\phi^{-2}$. We obtain the following theorem.
% Previous wording: This result is given in the preliminaries, Section~\ref{sec:preliminaries}, since most of it is known and is scattered among various papers~\cite{OstrovskyRSS12, AwasthiBS10, KumarK10}, and should come as no surprise considering it is known that all of these notions (and others) yield a PTAS for the clustering problem~\cite{Cohen-AddadS17} 
% \footnote{With a small additive error, as {\em private} $k$-means algorithms must have such error as well. See Theorem \ref{thm:main}.}

\begin{theorem}[informal]\label{them:intro1}
There exists an $(\eps,\delta)$-differentially private algorithm such that the following holds. 
Let $X$ be a database containing $n$ points in the $d$-dimensional ball $\BBB(0,1)$, and assume that $X$ is $\phi$-separated for $k$-means for $\phi=O(1)$ (sufficiently small). When applied to $X$, the algorithm returns (w.h.p.)\ a set of $k$ centers $C'$ satisfying
$\cost_X(C')\leq(1+O(\phi^2))\cdot\opt_k(X)+\Delta$, for
$\Delta\lesssim\frac{k(\sqrt{d}+\sqrt{k})}{\eps}$.\footnote{In the introduction, we use the informal notation $a\lesssim b$ to signify $a \leq b^{(1+\gamma)}\cdot \polylog(k,n,\eps,\delta^{-1})$ for some small constant $0<\gamma<<1$. Similarly for $\gtrsim$.}
\end{theorem}

To the best of our knowledge, this is the first differentially private algorithm that guarantees multiplicative error smaller than 2 in term of the $k$-means cost (for $\phi$-separated datasets, provided that $\phi$ is small enough). 
We analyze this algorithm's utility also in terms of Wasserstein distance, as follows.

\begin{theorem}[informal]\label{thm:intro2}
There exists an $(\eps,\delta)$-differentially private algorithm such that the following holds. 
Let $X$ be a database containing $n$ points in the $d$-dimensional ball $\BBB(0,1)$. Assume that $X$ is $\phi$-separated for $k$-means for $\phi=O(1)$ (sufficiently small), and assume that
$\opt_{k-1}(X)\gtrsim\frac{k(\sqrt{d}+\sqrt{k})}{\phi^2\cdot\eps}$.
%$$\opt_{k-1}(X)\geq\tilde{O}\left(\frac{k^{1.01}\cdot d^{0.51}\cdot\Lambda^2}{\eps^{1.01}}+\frac{k^{1.5}\cdot\Lambda^2}{\eps}+\frac{k^{0.51}\cdot d^{0.26}}{\phi^2\cdot\eps^{0.51}}+\frac{k^{0.75}}{\phi^2\cdot\eps}\right).$$ 
When applied to $X$, the algorithm returns (w.h.p.)\ a set of $k$ centers $C'$ satisfying $\wdist(C,C')\leq O(\phi^2)$, where $C$ are the optimal $k$-means centers.
\end{theorem}

The error bound in this theorem matches the state-of-the-art previous result of Huang and Liu~\cite{HuangL18}, and offers some improvements in terms of the requirement on $\opt_{k-1}(X)$.\footnote{Specifically, the bound of \cite{HuangL18} is guaranteed to hold whenever $\opt_{k-1}(X)\gtrsim n^{\frac{11}{20}}k^{\frac{7}{4}}d^{\frac{3}{4}}\eps^{-\frac{1}{2}}\phi^{-4}$, whereas our bound holds also for smaller values of $\opt_{k-1}(X)$.} %In particular, in our construction $\opt_{k-1}(X)$ dependency on $n$ is only poly-logarithmic.

\begin{table}[htbp]
\begin{adjustwidth}{-18pt}{-14pt}
\begin{footnotesize}
  \centering
  \caption{Algorithms for $k$-means with separable input}
    \begin{tabular}{|c|c|c|c|}
    \hline
      \multirow{2}{*}{\bf{Reference}}                        &  \multicolumn{2}{c|}{\bf{Wasserstein distance}}  & \multirow{2}{*}{\bf{${\boldsymbol{k}}$-means cost}} \\\cline{2-3}
    & ${\boldsymbol{D_w(C',C)}}$ & \bf{Requirement on} ${\boldsymbol{n}}$ & \\ \hline\hline
    Nissim et al.~\cite{NRS07}       & 
    $O(\epsilon^{-1} d \sqrt{k} \phi^2)$ & $\Omega(k^{\frac{1}{3}} d^{\frac{1}{3}} \phi^{-\frac{4}{3}} )$
      &  --    \\\hline
    %Wang et al. \cite{Wang2015}  2015    & $O(\phi^2 \cdot \sqrt{k} )$  & -   & - \\
    Huang and Liu \cite{HuangL18}  & 
    $O(\phi^2)$  & $\Omega(k^{3.88} d^{1.66} \eps^{-1.11} \phi^{-8.88})$
     &  --   \\
    \hline
    This work (Section~\ref{sec:main})                           & $O(\phi^{2})$      &
     $\tilde{\Omega} (k^{1.51} d^{0.51} \eps^{-1.01} \phi^{-2})$
    & $(1+O(\phi^2))\cdot \opt_k(X) + \tilde{O} \left( \frac{k(\sqrt{d}+\sqrt{k})}{\eps} \right) $ \\\hline
    This work (Section~\ref{sec:alternative})                     & $O(\phi^2)$   &      $ \tilde{\Omega} (k^{3.51}d^{1.51}\eps^{-1}\phi^{-2} )$                  & $\left(1+O(\phi^2)\right)\cdot\opt_k(X) + \tilde{O}\left( \frac{k \sqrt{d}}{\eps}\right) $ \\
    \hline
    \end{tabular}%
    \begin{tablenotes}
      \small
      \item Where $C$ is the optimal center set of $X$, $C'$ is the center set returned from the algorithm and $\opt_k(X)$ is the optimal $k$-means cost of $X$. We use the $\tilde{O}$ and $\tilde{\Omega}$ notation for disregarding logarithmic factors.
    \end{tablenotes}
\end{footnotesize}
\end{adjustwidth}
\end{table}

Due to the simplicity of our algorithm, we give its local-model analogue and $k$-median analogue in Sections~\ref{sec:kMeansStabilityLocal} and~\ref{sec:median}, respectively. 
This is the first locally-private algorithm for clustering stable instances (even w.r.t.\ the Wasserstein distance). Unlike our algorithm, the previous constructions of~\cite{NRS07,Wang2015,HuangL18} are based on the {\em sample-and-aggregate} framework\footnote{Sample-and-aggregate framework (Nissim, Raskhodnikova and Smith 2007 \cite{NRS07}) is a general method for privately compute a function over an input, typically used for function with high global sensitivity. The stages of the method are first sample the input into $T$ samples each of size $m$, then compute (non privately) a function over each sample $f_i(s_i)$, then, aggregation stage, privately compute an output $g(\{f_i(s_i)\}_{i\in[m]}, \eps, \delta)$.}~\cite{NRS07}, which is inapplicable (in general) in the local-model. It is hence unclear whether the constructions of~\cite{NRS07,Wang2015,HuangL18} have analogues for the local-model.

Lastly, we revise the sample-and-aggregate based approach of Huang and Liu by replacing their aggregation stage with a similar Lloyd step. We then iterate through several of the claims made by Huang and Liu and improve upon their bounds to achieve better utility guarantees. The improved sample-and-aggregate based algorithm and its analysis is given in Section~\ref{sec:alternative}.\\

This thesis is based on a joint work with Or Sheffet and Uri Stemmer (AISTATS 2020 \cite{shechner2020private}).
\newpage

%%%%%%%%%%%%%%%%%%%%%%
% Related Literature
%%%%%%%%%%%%%%%%%%%%%%
\section{Related Literature}

\subsection{Prior work on center based clustering}
Even without privacy constraints, the $k$-means and $k$-median problems are NP-hard.
Their most simple and known heuristic is the \text{`}Lloyd-Forgey method\text{'}. In 1957 Lloyd presented a heuristic algorithm for the discrete $k$-means problem (published in 1982 \cite{lloyd1982least}). In 1965 Forgey gave similar heuristic algorithm to the continuous $k$-means problem (see formulation in \cite{forgey1965cluster}). 
The method is an expectation maximization iterative method and can be used for $k$-median as well \cite{bradley1997clustering} (with some modifications), but as both problem are not convex, it has no guarantee to converge to the general optimum. In addition, although the run time till a convergence is achieved is usually fast, for $k$-means there are instances of 2 dimensions that require exponential time of $2^{\Omega(n)}$ to converge \cite{DBLP:journals/dcg/Vattani11}.
% Checked: the lloyd forgy method can be adapted for k-means as well. in the cited paper the writers show a translation from the linear programming form of the problem to other linear programming problem. From brief readding it seams that the translated program is solved to find the 'maximization' stage - the median. in the Wikipedia it states that the k-median are found by finding a median in each dimantion seperatley.
% Expectation maximization - preperation for a footnote if need be: that is an iterative method for optimzing a probability distribution parameters, given data that is assumed to be drawn from that probability. The search is done on likelihood function of the parameters given the data. For a multimodal likelihood function (with more then one maxima) the method has no guarantee to converge to the global optimum, rather to a local one. Expectation stage: (naming is somewhat misleading) is a stage loss function update with previous parameters, and data. maximization stage is optimizing parameters for the updated loss function, w.r.t the likelihood function.

%%%%%%%%%%%%%%%%%%%%%%%%%%%%%%%%%%%%%%%%%%
% Hardness results results:
%%%%%%%%%%%%%%%%%%%%%%%%%%%%%%%%%%%%%%%%%%
Finding exact (optimal) solution both to the Euclidean $k$-means problem and to the Euclidean $k$-median problem is NP-hard (even on very basic settings)~\cite{aloise2009np,dasgupta2009random,drineas2004clustering,DBLP:journals/tcs/MahajanNV12,megiddo1984complexity}. Furthermore, approximating the $k$-means and the $k$-median problems is NP-hard as well: First, Jain et al.~\cite{jain2002new} showed in 2002 that unless $NP \subseteq DTIME[n^{O(\log{\log{n}})}]$, it is hard to approximate the $k$-median to within a factor of $1+2/e \approx 1.735$, and the $k$-means to within a factor of $1+ 8/e \approx 3.943$ (see also \cite{ahmadian2019better}).
Then, in 2015, Awasthi et al.~\cite{DBLP:journals/corr/AwasthiCKS15} showed that there exist a constant $\alpha>0$ s.t.\ it is NP-hard to approximate the Euclidean $k$-means up to $(1+\alpha)$. Finally, in 2017, Lee et al.~\cite{lee2017improved} showed that $\alpha \geq 0.0013$.

%%%%%%%%%%%%%%%%%%%%%%%%%%%%%%%%%%%%%%%%%%
% Positive results: approximation schemes
%%%%%%%%%%%%%%%%%%%%%%%%%%%%%%%%%%%%%%%%%%
Due to the hardness results mentioned, a lot of effort was invested on constructing efficient algorithms with guaranteed approximation bound for both $k$-means and $k$-median.
We first survey some of the results for $k$-means. In 2004 Kanungo et al.~\cite{kanungo2004local} showed a $(9+\alpha)$-multiplicative factor guarantee based on local search heuristic - method of successively improving a set of centers by swapping one of them with an improving candidate. They also showed a nearly tight $(9-\alpha)$-multiplicative factor lower bound for any approach based on performing fixed number of swaps, therefore ended the effort of finding a better constant approximation guarantee algorithm based on local search. Best constant-factor approximation algorithm to date achieves multiplicative factor guarantee of $6.357$ (Ahmadian et al.\ 2016 \cite{ahmadian2019better}). A PTAS was given for a case where the problem parameter $k$ is fixed. First by Inaba et al.\ (1994 \cite{Inaba94}) for $k=2$, achieving $(1+\alpha)$-multiplicative factor guarantee for any fixed $\alpha$ with running time of $O(n\alpha^{-d})$. Later, Kumar et al.\ (2004 \cite{kumar2004simple}) and then Feldman et al.\ (2007 \cite{feldman2007ptas}) showed similar results with running time of $O(2^{(k/\alpha)^{O(1)}}nd)=O(nd)$ and $O(nkd+d\cdot poly(k/\alpha) + 2^{\tilde{O}(k/\alpha)}) = O(nd)$ respectively.

We next survey some of the results for $k$-median. 
Before a constant approximation algorithm was found, a PTAS was presented by Arora et al.\ in 1998 \cite{arora1998approximation} to the planar ($d=2$) Euclidean $k$-median. In 1999 Charikar et al.~\cite{charikar2002constant} presented the first constant-factor approximation algorithm for the metric $k$-median problem with approximation factor of $6\frac{2}{3}$. Best constant-factor approximation algorithm is due to Ahmadian et al.\ (2016 \cite{ahmadian2019better}) achieving multiplicative factor guarantee of $\approx 2.633$ for the Euclidean $k$-median problem. 
Current approximation algorithms for $k$-means and $k$-median are summarized in Tables~\ref{table1} and~\ref{table2}.

%%%%%%%%%%%%%%%%%%%%%%%%%%%%%%%%%%
% $k$-means Recent Results table
%%%%%%%%%%%%%%%%%%%%%%%%%%%%%%%%%%
\begin{table}[htbp]
\begin{adjustwidth}{-15pt}{-15pt}
  \centering
  \caption{Recent $k$-means clustering results}\label{table1}
    \begin{tabular}{|c|c|c|}
    %\toprule
    \hline
    \bf{Reference} &\bf{Multiplicative Error} & \bf{Run Time (For PTAS)} \\
    \hline\hline
    %\midrule
    Kanungo et al.\ (2004) \cite{kanungo2004local}& $(9+\alpha)$     &  $\poly(n,k,d,\alpha^{-1})$ \\
    \hline
    %\midrule
    Kumar et al.\ (2004) \cite{kumar2004simple}& $(1+\alpha)$-PTAS     & $O(2^{(k/\alpha)^{O(1)}}\cdot nd)$ \\
    \hline
    %\midrule
    Feldman et al.\ (2007) \cite{feldman2007ptas}& $(1+\alpha)$-PTAS     &  $O(nkd+d\cdot \poly(k/\alpha)+2^{\tilde{O}(k/\alpha)})$\\
    \hline
    %\midrule
    Ahmadian et al.\ (2016) \cite{ahmadian2019better} & $6.357$ &  $\poly(n,k,d)$ \\
    \hline
    %\bottomrule
    \end{tabular}%
\end{adjustwidth}
\end{table}%

%%%%%%%%%%%%%%%%%%%%%%%%%%%%%%%%%%
% $k$-median Recent Results table
%%%%%%%%%%%%%%%%%%%%%%%%%%%%%%%%%%
\begin{table}[htbp]
  \centering
  \caption{Recent $k$-median clustering results}\label{table2}
    \begin{tabular}{|c|c|c|}
    %\toprule
    \hline
    \bf{Reference} & \bf{Multiplicative Error} & \bf{Run Time (For PTAS)} \\
    \hline\hline
    %\midrule
    Arora et al.\ (1998) \cite{arora1998approximation}& $(1+\alpha)$-PTAS     &  $O(n^{O(1+1/\alpha)})$  \\
    \hline
    %\midrule
    Charikar et al.\ (1999) \cite{charikar2002constant}& $6\frac{2}{3}$     &  $\poly(n,k,d)$ \\
    \hline
    %\midrule
    Ahmadian et al.\ (2016) \cite{ahmadian2019better}& $2.633$     & $\poly(n,k,d)$ \\
    \hline
    %\bottomrule
    \end{tabular}%
\end{table}%

\subsection{Prior work on private center based clustering}
Our interest lies in {\em private} algorithms for approximating $k$-means and $k$-median problems. By now there has been a lot of work on private $k$-means and $k$-median algorithms, which resulted in all cases in additive error in addition to the multiplicative error. This is in contrast to the non private results which had only multiplicative error. Specifically, Gupta et al.~\cite{DBLP:conf/soda/GuptaLMRT10} showed that introducing privacy constraints for the $k$-median problem must result in an additive (at least $\Omega(\Lambda \cdot \ln(n/k)/\eps )$) error. Similarly, one can show that every differentially private algorithm for $k$-means must have additive error $\Omega(\Lambda^2)$ (see, e.g.,~\cite{DBLP:journals/corr/abs-1804-08001}). 
Recall that $\Lambda$ is the diameter of the instance (i.e.\ all points in $X$ are taken from a ball centered at the origin with a diameter of $\Lambda$).

We next survey the works on private clustering, which are most relevant for this thesis. Gupta et al.~\cite{DBLP:conf/soda/GuptaLMRT10} showed in 2010 an $\eps$-differentially private algorithm for $k$-means and $k$-median, based on the classical local search heuristic, with a good approximation guarantee ($6\cdot \opt + O(\Lambda \cdot k^2 \cdot \log n /\eps )$), but it had an exponential run time. Balcan et al.~\cite{DBLP:conf/icml/BalcanDLMZ17} in 2017 applied Gupta's \cite{DBLP:conf/soda/GuptaLMRT10} approach, while the centers were searched within a small set of center candidates. That lead to an efficient algorithm but deteriorated the approximation guarantee. 
%TBR: below line of work was done for k-means, But it can be also transformed to k-median. how to phrase that? and why?
%It looks like because the centers of both k-means and k-median reside IN the enclosing ball, therefore the algorithm is the same for both problems.\\
Different approach was shown by Feldman et al.~\cite{DBLP:conf/ipsn/FeldmanXZR17} in 2017, which iteratively searches smallest ball that enclose $t$ points of the data, and setting a center for the found points while eliminating them from the next iteration search. An improved algorithm was then shown by Nissim and Stemmer \cite{DBLP:conf/alt/NissimS18} to reach a multiplicative error of $O(k)$. 
State-of-the-art differentially private $k$-means clustering  achieves a multiplicative error of $O(1)$ by Kaplan and Stemmer  \cite{DBLP:journals/corr/abs-1804-08001}. We summarize these results in Table~\ref{table4}.

\begin{table}[htbp]
  \centering
  \caption{Recent private $k$-means clustering results}\label{table4}
    \begin{tabular}{|c|c|c|}
    %\toprule
    \hline
    \bf{Reference} & \bf{Multiplicative Error} & \bf{Additive Error} \\
    \hline\hline
    %\midrule
    Balcan et al.\ (2017) \cite{DBLP:conf/icml/BalcanDLMZ17}& $O(\log^3 n)$     &  $\tilde{O}\left(\frac{d+k^{2}\eps}{\eps^{2}}\Lambda^2\right)$\\
    \hline
    %\midrule
    Feldman et al.\ (2017) \cite{DBLP:conf/ipsn/FeldmanXZR17}& $O(k\log n)$     & $\tilde{O}\left(\frac{\sqrt{d}k^{1.5}}{\eps}\Lambda^2\right)$ \\
    \hline
    %\midrule
    Nissim and Stemmer (2018) \cite{DBLP:conf/alt/NissimS18}& $O(k)$     &  $\tilde{O}\left(d^{0.51}k^{1.51}\Lambda^2\right)$\\
    \hline
    %\midrule
    Kaplan and Stemmer (2018) \cite{DBLP:journals/corr/abs-1804-08001} & $O(1)$ & $\tilde{O}\left(\left( d^{0.51}k^{1.01}+k^{1.51}\right)\Lambda^2\right)$ \\
    \hline
    %\bottomrule
    \end{tabular}%
\end{table}%

\subsection{Prior work on clustering stable instances (non-privately)}
The line of work on clustering stable (or ``nice'') instances was motivated by the gap between the success of the simple `Lloyd-Forgy' method over real data and the theoretical hardness of the problem. While performing well on real data, this method had no convergence guarantees on several axis: First, the method had no guarantee to converge to the global optimum. Second, the existence of instances that converging to the global optimum require exponential time. Third, the relatively easy construction of instances that can converge to a {\em local} optimum that is {\em arbitrary} far from the global optimum (e.g.\ see \cite{kanungo2004local}). This gap lead to the following question: 
\begin{question}
Are ``real'' instances in fact easier than worst case instances? What are the properties of ``real'' instances?
\end{question}

%%%%%%%%%%%%%%%%%%%%%%%%%%%%%%%%%%%%%%%%%%%%%%
% Seperation Criteria
%%%%%%%%%%%%%%%%%%%%%%%%%%%%%%%%%%%%%%%%%%%%%%
\textbf{Clustering beyond worst case analysis:} Instead of constructing private clustering algorithms which are suppose to operate well on every possible input, we want to design algorithms which are only guaranteed to work well when their inputs are ``nice". The upside is that we can aim for much higher accuracy guarantees. One such possible definition for ``niceness" is the following, due to Ostrovsky et al.~\cite{DBLP:journals/jacm/OstrovskyRSS12}.

\begin{definition}[$\phi$-Separability input criteria] \label{def:seperationCriteria}
A clustering instance $X$ is {\em $\phi$-well-separated} for 
$k$-clustering objective with parameter $p$ if the following holds:
$$ \frac{\plainopt^p_k(X)}{\plainopt^p_{k-1}(X)} \leq \phi ^p  $$ 
\end{definition}
Above criteria requires an upper bound on the ratio between the optimum values of the $k$-clustering and $(k-1)$-clustering objective functions. Seemingly a simple quantity, it in fact implies strong geometrical features on the instance (see Theorem~\ref{thm:ostKmeans}). It is also a particular case of other more recently defined criteria (e.g.\ Definition~\ref{def:stabNotionABProperty}, Definition~\ref{def:stabNotionProxCond}), as we survey next.

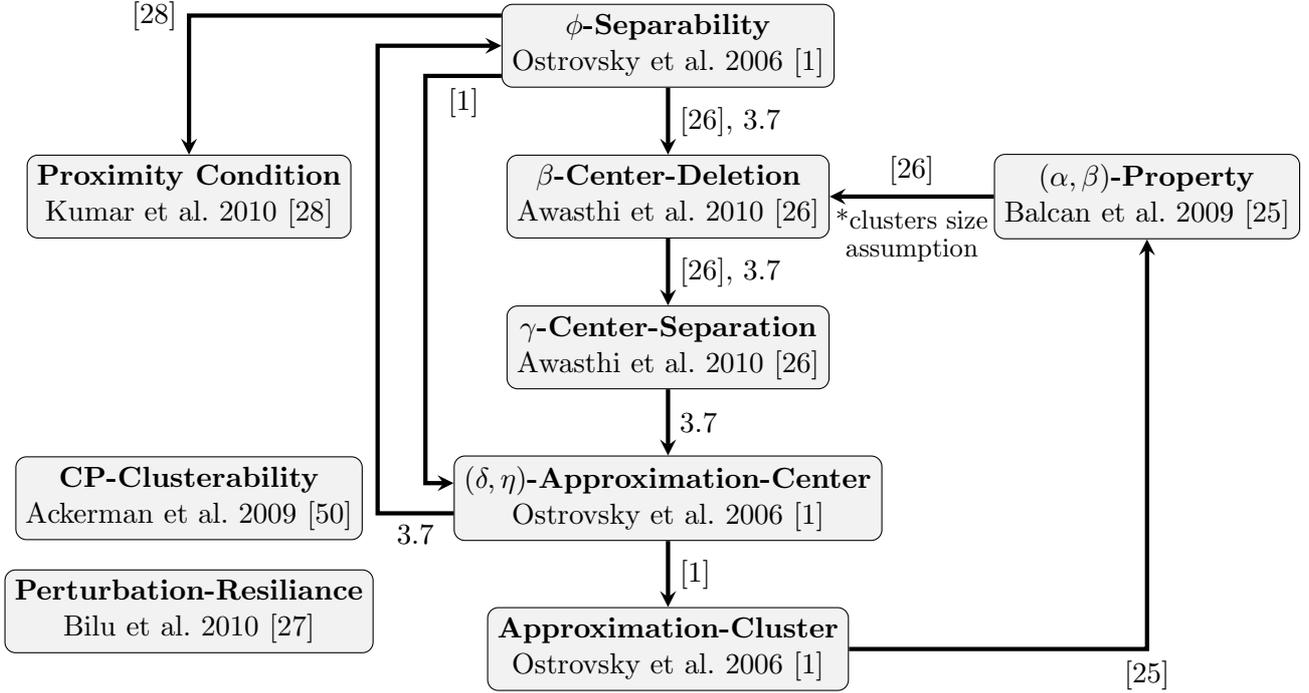
\begin{figure}[htb]
\centering
\caption{Stability notions implication flow chart} \label{fig:StabilityNotionFlowchart}

\makebox[0pt][c]{
\begin{small}
\begin{tikzpicture}[node distance=2cm]
\node (ORSS1) [stability_assumption, align=center] 
{ \textbf{$\phi$-Separability} \\ 
Ostrovsky et al.\ 2006 \cite{DBLP:journals/jacm/OstrovskyRSS12}};

\node (ABS1) [stability_assumption, align=center, below of=ORSS1, yshift=-0.0cm] 
{\textbf{$\beta$-Center-Deletion} \\ 
Awasthi et al.\ 2010 \cite{AwasthiBS10}};

\node (ABS2) [stability_assumption, align=center, below of=ABS1, yshift=-0.0cm] 
{\textbf{$\gamma$-Center-Separation} \\ 
Awasthi et al.\ 2010 \cite{AwasthiBS10}};

\node (ORSS2) [stability_assumption, align=center, below of=ABS2, yshift=-0.0cm] 
{\textbf{$(\delta,\eta)$-Approximation-Center} \\ 
Ostrovsky et al.\ 2006 \cite{DBLP:journals/jacm/OstrovskyRSS12}};

\node (ORSS3) [stability_assumption, align=center, below of=ORSS2, yshift=-0.0cm] 
{\textbf{Approximation-Cluster} \\ 
Ostrovsky et al.\ 2006 \cite{DBLP:journals/jacm/OstrovskyRSS12}};

\node (BBG) [stability_assumption, align=center, right of=ABS1, xshift=4.3cm] 
{\textbf{$(\alpha,\beta)$-Property} \\ 
Balcan et al.\ 2009 \cite{BalcanBG09}};

\node (KK) [stability_assumption, align=center, left of=ABS1, xshift=-4.3cm] 
{\textbf{Proximity Condition} \\ 
Kumar et al.\ 2010 \cite{KumarK10}};

\node (BL) [stability_assumption, align=center, below of=KK, yshift=-3.5cm] 
{\textbf{Perturbation-Resiliance} \\ 
Bilu et al.\ 2010 \cite{BiluL10}};

\node (AB) [stability_assumption, align=center, below of=KK, yshift=-2.0cm] 
{\textbf{CP-Clusterability} \\ 
Ackerman et al.\ 2009 \cite{AckermanB09}};

% arrows of implications
\draw [implication] (ORSS1) -- 
node[right] {\cite{AwasthiBS10}, \ref{lem:StabilityNotionsEquivalence}} (ABS1);

\draw [implication] ([yshift=-4mm]ORSS1.west) -- 
node[below] {\cite{DBLP:journals/jacm/OstrovskyRSS12}} ++(-1cm,0) |- 
([yshift=2mm]ORSS2.west) ;

\draw [implication] (ABS1) -- 
node[right] {\cite{AwasthiBS10}, \ref{lem:StabilityNotionsEquivalence}} (ABS2);
\draw [implication] (ABS2) -- 
node[right] {\ref{lem:StabilityNotionsEquivalence}}(ORSS2);
\draw [implication] (ORSS2) -- 
node[right] {\cite{DBLP:journals/jacm/OstrovskyRSS12}} (ORSS3);
\draw [implication] ([yshift=-2mm]ORSS2.west) -- 
node[below] {\ref{lem:StabilityNotionsEquivalence}}++(-1cm,0) |- 
([yshift=0mm]ORSS1.west) ;

%\draw [implication] ([xshift=1mm] ORSS2.north west) |- node[left] {\ref{lem:StabilityNotionsEquivalence}} (ORSS1.west);
\draw [implication] (ORSS3.east) -| 
node[below] {\cite{BalcanBG09}} (BBG.south);
\draw [implication_weak] (BBG.west) -- 
node[above] {\cite{AwasthiBS10}}
node[below] {{\footnotesize *clusters size}} 
node[below, yshift=-0.4cm] {{\footnotesize assumption}}
 (ABS1.east);

\draw [implication] ([yshift=4mm] ORSS1.west) -| 
node[left] {\cite{KumarK10}} (KK);

\end{tikzpicture}
\end{small}
}
\fnote{An arrow between stability notions indicates an implication. For example, if a dataset satisfies $\phi$-Separability then it also satisfies Proximity Condition, i.e., $\phi$-Separability is a stronger assumption. The implications references are adjacent to the corresponding arrow.}
\end{figure}

\paragraph{Notions of clustering instance stability:} 
The line of work on clustering stable instances has produced many interesting results, including several different notions of input stability. \cite{BalcanBG09, BiluL10, AwasthiBS10, KumarK10, AwasthiBS12}
See \cite{AckermanB09} for a survey of some of these notions. We now present several such stability notions, and illustrate the relations between them (see Figure~\ref{fig:StabilityNotionFlowchart}).

%%%%%%%%%%%%%%%%%%%%%%%%%%%%%%%%%%%%%%%%%%%%%%
% Center Perturbation Clusterability
%%%%%%%%%%%%%%%%%%%%%%%%%%%%%%%%%%%%%%%%%%%%%%
\paragraph{•}\textbf{Center perturbation clusterablity.}
Two sets of centers are said to be {\em $\alpha$-close} if for each center in one set there is a center in the other that reside within a radius of $\alpha$ from it. A clustering instance $X$ is said to be {\em center perturbation clusterable} if every set of centers which is $\alpha$-close to the optimal set of centers also have low clustering cost. Formally,

\begin{definition}[Center Perturbation Clusterability \cite{AckermanB09}] \label{def:stabNotionCenterPertStab} A data set $X$ is $(\alpha, \beta)$-CP clusterable for $k$-means (for $\alpha, \beta \geq 0$) if for every center set $C'$ that is $\alpha$-close to some optimal center set of $X$ the following hold:
$$\cost_{X}(C')\leq (1+\beta)\cdot \opt_{k}(X)$$ 
Where two center sets $C,C'\in (\R^d)^k$ are said to be $\alpha$-close if there exist some permutation $\sigma\in \Pi_{k}$ s.t.\ $\forall i\in [k]. \|c_i - {c'}_{\sigma(i)}\|\leq \alpha$
\end{definition}

Ackerman and Ben-David \cite{AckermanB09} presented an algorithm that returns a set of $k$-centers whose cost is within $(1+\beta)$ multiplicative factor of the optimum for $k$-means and $k$-median, provided that the instance is $( \Lambda/\sqrt{d}, \beta )$-CP clusterable.

%%%%%%%%%%%%%%%%%%%%%%%%%%%%%%%%%%%%%%%%%%%%%%
% $\beta$-center-deletion
%%%%%%%%%%%%%%%%%%%%%%%%%%%%%%%%%%%%%%%%%%%%%%
\paragraph{•}\textbf{$\beta$-Center-Deletion} is a stability notion defined by \cite{AwasthiBS10} (where it is called `{\em $\beta$-weak-deletion stability}'). This notion, as $\phi$-separability, requires a bound on the relation between $\opt_k(X)$ to the cost of some $(k-1)$-clustering, but not necessarily to the optimal one. Formally:
\begin{definition}[$\beta$-Center-Deletion \cite{AwasthiBS10}]\label{def:stabNotionCenterDeletionStab} A clustering instance is said to be $\beta$-Center-Deletion stable for $\beta > 1$ if for every cluster $i$ from the optimal clustering, deleting its center and assigning all of its points to any center $c_j, j\neq i$ results in a $(k-1)$-clustering with a cost $\geq \beta\cdot \opt_k$
\end{definition}
Observe that if an instance $X$ is $\phi$-separable, then it is also, in particular, $\phi^{-p}$-Center-Deletion stable.
Awasthi et al.~\cite{AwasthiBS10} also defined the following stability notion.

%%%%%%%%%%%%%%%%%%%%%%%%%%%%%%%%%%%%%%%%%%%%%%
% $\gamma$-Center-Separation
%%%%%%%%%%%%%%%%%%%%%%%%%%%%%%%%%%%%%%%%%%%%%%
\paragraph{•}\textbf{$\gamma$-Center-Separation} is a stability notion  that binds the size of each cluster in the optimal clustering and the distance from its center to the closest adjacent center. Formally:
\begin{definition}[$\gamma$-Center-Separation  \cite{AwasthiBS10}] \label{def:stabNotionCenterSeparationStab}
Let $X_i\subseteq X$ be the cluster $i$, and denote its size by $|X_i|$. Let $D^2_i=\min_{j\neq i}{\|c_i-c_j\|^2}$ be the distance between the center $c_i$ and its closest center $c_j$ from the optimal $k$-center-set. A $k$-means instance is said to be $\gamma$-center-separation stable for some $\gamma>0$ if for all $i\in [k]$ the following holds:
$$D^2_i\geq \frac{\gamma}{|X_i|}\cdot \opt_k$$
\end{definition}

%%%%%%%%%%%%%%%%%%%%%%%%%%%%%%%%%%%%%%%%%%%%%%
% $(\delta, \frac{1}{4})$-Approximation-Center Stability
%%%%%%%%%%%%%%%%%%%%%%%%%%%%%%%%%%%%%%%%%%%%%%
\paragraph{•}\textbf{$(\delta, \eta)$-Approximation-Center Stability} is a stability notion that considers an instance a stable one if all of near optimal clusterings in term of cost, have their centers similar to the optimal centers (correspondingly by proximity). It states that an instance is $(\delta, \eta)$-Approximation-Center stable if for all its clustering that are $\delta$-multiplicative factor away from the optimal cost, their centers have clear adjacency to the optimal centers (one to one). Formally:
\begin{definition}[$(\delta, \eta)$-Approximation-Center Stability \cite{DBLP:journals/jacm/OstrovskyRSS12}] \label{def:stabNotionApproximationCenterStab} Denote $(c_1,...,c_k)$ the optimal set of centers for a data set $X$ for $k$-means, let $\delta>1, \eta<\frac{1}{2}$ and let $D^2_i=\min_{j\neq i}{\|c_i-c_j\|^2}$. Then $X$ is $(\delta,\eta)$-Approximation-Center stable for $k$-means if for any $k$-tuple 
$\hat{c_1},...,\hat{c_k}$ of cost at most $\delta\cdot \opt_k$ we have a matching $\varphi$ s.t.\ for all $i\in [k]$ the following holds:
$$\|c_i - \hat{c_{\varphi(i)}}\|^2 < \eta\cdot D^2_i $$

\end{definition}
This notion has a useful geometrical property allowing to bound the error of near optimal set of centers; indeed we use that property in this work. 

%%%%%%%%%%%%%%%%%%%%%%%%%%%%%%%%%%%%%%%%%%%%%%
% Approximation-Center Stability
%%%%%%%%%%%%%%%%%%%%%%%%%%%%%%%%%%%%%%%%%%%%%%
%\textbf{Approximation-Cluster Stability} is a stability notion that have the same spirit as the previous, and is implied by it. It states that an instance is Approximation-Cluster stable if all its clustering with a near optimal cost agree on most of the points, meaning that essentially they are similar. Formally:
%\begin{definition}[title]
%Let $X\substeq \R^d$ be a dataset and its partition $\{X_1,...,X_k\}$ be its optimal clustering for $k$-means where $\{c_1,...,c_k\}$ are the clusters centers. Denote $\{\hat{c}_1,...,\hat{c}_k\}$ as a set of centers and their clustering of $X$ to be $\hat{C}_1,...,\hat{C}_k$. Finally let $0<\alpha<\frac{1-401\epsilon^2}{400}$. Then $X$ is {\em Approximation-Cluster} stable if 
%\end{definition}

%%%%%%%%%%%%%%%%%%%%%%%%%%%%%%%%%%%%%%%%%%%%%%
% $(\alpha, \beta)$-Property
%%%%%%%%%%%%%%%%%%%%%%%%%%%%%%%%%%%%%%%%%%%%%%
\paragraph{•}\textbf{$(\alpha, \beta)$-Property} is a stability notion that focuses on the {\em clusterings} explicitly (rather than focusing on center sets, or focusing on clusterings that are induced from center sets). Namely, a clustering $C$ is a partition of the input $X$: $C=(C_1,...,C_k)$ s.t.\ $\uplus_{i\in [k]}{ \{ C_{i}\} } = X$. An instance have $(\alpha, \beta)$-Property if all its clusterings with a cost within a multiplicative factor $\alpha$ from the optimal cost, have a bounded symmetric difference w.r.t the optimal clustering. Formally:    
\begin{definition}[$(\alpha, \beta)$-Property \cite{BalcanBG09}] \label{def:stabNotionABProperty}A data set $X$ has the $(\alpha, \beta)$-Property for $k$-clustering (for $\alpha \geq 1, \beta \geq 0$) if for every clustering $C'$ (with corresponding clusters means $c'=(c_1',...,c_k')$) that is $\beta$-close to the optimal clustering $X$ the following hold:
$$\cost_{X}(c')\leq \alpha\cdot \opt_{k}(X)$$ 
Where two clusterings $C,C'\in (\R^d)^k$ are said to be $\beta$-close if there exist some permutation $\sigma\in \Pi_{k}$ s.t.\ $\frac{1}{n}\sum_{i\in[k]}{|C_i \triangle {C'}_{\sigma(i)}|} \leq \beta$ (and $\triangle$ denotes the symmetric difference).
\end{definition}
Ostrovsky et al.~\cite{DBLP:journals/jacm/OstrovskyRSS12} showed that $\phi$-separated instance admit a similar but stronger property than $(\alpha, \beta)$-Property. Namely, for any $\phi$-separated instance, any clustering with cost with a multiplicative factor of $\gamma\cdot \phi^2$ from the optimal cost has {\em per-cluster} bounded symmetric difference of $O(\phi^2\cdot  |C_i|)$ from the corresponding optimal cluster $X_i$. In particular, such an instance also satisfies the $(\gamma\cdot \phi^2, O(\phi^2))$-Property. The converse is not necessarily true~\cite{BalcanBG09}.\\

%%%%%%%%%%%%%%%%%%%%%%%%%%%%%%%%%%%%%%%%%%%%%%
% stability notions implications flowchart
%%%%%%%%%%%%%%%%%%%%%%%%%%%%%%%%%%%%%%%%%%%%%%
% centering the chart flow

\paragraph{•}\textbf{Perturbation resilience} is a stability notion that quantizes the amount of distortion permitted to the space {\em metric} without changing the optimal clustering. Formally: %, rather than stating  some condition on the instance clusterings.
\begin{definition} [Perturbation resilience \cite{BiluL10, AwasthiBS12}] Given an instance $(X,d)$ where $X$ is set of $n$ points and $d$ is a metric for $X$, then we call $(X,d)$ $\alpha$-perturbation resilient for $k$-clustering under $d$ if for any $\alpha$-perturbation of $d$, say $d'$, the (only) optimal clustering of $(X,d')$ is identical to the optimal clustering of $(X,d)$.\\
Given a metric $(X,d)$ and $\alpha > 1$, we say a function $d':X \times X \rightarrow \R_{\geq0}$ is an {\em $\alpha$-perturbation} of $d$ if $\forall x_1, x_2 \in X. d(x_1,x_2)\leq d'(x_1,x_2) \leq \alpha d(x_1,x_2)$. There is no requirement from $(X,d')$ to be a metric. 
\end{definition}
Awasthi et al.~\cite{AwasthiBS12} presented an algorithm that finds the optimal clustering of $\alpha$-perturbation resilient instances, for any $\alpha \geq 2+\sqrt{3}$. Balcan and Liang \cite{balcan2016clustering} presented an algorithm that relaxes the requirement on the perturbation constant $\alpha$ for any $\alpha \geq 1+\sqrt{2}$. The result holds for any center based clustering objective, therefore it holds for $k$-means and $k$-median.

\paragraph{•}\textbf{Proximity condition} is a condition classifying a clustering instance $X$'s points into two sets: `good' set denoted by $G\subseteq X$ and the rest of the points. For the clustering problem with optimal center set $(c_1,...,c_k)$, {\em Proximity condition} for a point $x\in X_i$ holds if it is closer to its center $c_i$ then to any other center $c_j$ by at least $\Delta_{i,j}$ - a quantity defined by the {\em spectral norm} of a centered version of the input matrix. Formally:

\begin{definition} [Proximity condition \cite{KumarK10}] \label{def:stabNotionProxCond} We are given $n$ points in $\R^d$ divided into $k$ clusters, namely $X_1,...,X_k$. Denote for $i\in [k]$ the mean of $X_i$ as $c_i$ and $n_i=|X_i|$. Let $X$ be a matrix $n\times d$ with rows corresponding to the points. Let $C$ be a matrix $n \times d$ where $C_j=c_i$ for all $j\in X_i$. We say a point $x\in X_i$ satisfy the proximity condition if for any $j\neq i$, the projection of  $x$ onto the $(c_i, c_j)$ line is at least $\Delta_{i,j}$ closer to $\mu_i$ than to $\mu_j$,
and the quantity $\Delta_{i,j}$ defined by:
$$\Delta_{i,j}=p\cdot k\left( \frac{1}{\sqrt{n_i}}+\frac{1}{\sqrt{n_j}} \right)\cdot \|X-C\|$$
Where $\|\text{ }\|$ is the spectral norm and $p$ is large enough constant.
\end{definition}

This notion is inspired by a related line of work: learning a {\em mixture of $k$ Gaussians}\footnote{Mixture of $k$ Gaussians: is a mixture distribution of $k$ Gaussian distributions i.e.\ the probability distribution derived from the following process: first an index $i$ is drawn from a categorical distribution with some $k$ weights, then the corresponding Gaussian distribution parameters are selected from $\{\langle \mu_i, \sigma_i \rangle\}_{i\in [k]}$ to draw the the data point from $N(\mu_{i}, \sigma^2_i)$.} distribution. This notion was defined by Kumar and Kannan in their paper~\cite{KumarK10}. In that paper they show that for a $k$-means instance with $|G|\geq (1-\alpha)\cdot |X|$ it is possible to correctly classify all but $O(\alpha \cdot k^2 |X|)$ of the points in polynomial time. Note that {\em proximity condition} does not relate to a quantity that measures per-cluster variance (as done on the line of work of learning mixture of Gaussians), rather to the variance (spectral norm) of the centered data. Kumar and Kannan also showed~\cite{KumarK10} that Proximity condition is implied by $\phi$-separability notion defined by Ostrovsky et al.~\cite{DBLP:journals/jacm/OstrovskyRSS12}. Namely, a $\phi$-separated instance admit the Proximity condition for all but a $\phi^2$-fraction of its points.

\subsubsection{Milestone results for clustering with stability assumptions}
The seminal work of Ostrovsky et al.~\cite{DBLP:journals/jacm/OstrovskyRSS12} presented the following  (non-private) results, under the stability assumption of Definition~\ref{def:seperationCriteria} ($\phi$-separability): 

\begin{enumerate}
\item A $(1+O(\phi^2))$-multiplicative approximation guarantee with success probability of $1-O(\sqrt{\phi})$ that runs in linear ($O(nkd+k^3d)$) time. \item A randomized PTAS algorithm with constant success probability and with an approximation factor of $(1+\alpha)$ and running time of $O(2^{O(k(1+\phi^2)/\alpha)}\cdot n\cdot d)$.
\end{enumerate} 
Note that the first result couples between the separability parameter $\phi$ and the approximation factor and success probability guarantees, and the second is for the case where the parameter $k$ is fixed.
Awasthi et al.~\cite{AwasthiBS10} showed deterministic PTAS algorithm with approximation factor of $(1+\alpha)$ that decouples the relation between the input separation parameter $\phi$ and the approximation factor, with running time polynomial in $(n,k)$ and exponential in $(1/\alpha, O(\phi^2))$.

\subsection{Other related work}
\paragraph{Center based clustering.} In addition to $k$-means and $k$-median, another center based clustering problem of interest is $k$-centers. In the $k$-centers problem we seek to find a set of $k$ centers that minimize the {\em maximum distance} from the input points to the centers. Formally:
\begin{definition}[$k$-centers cost function]
For center set $C\in (\R^d)^k$, $C=(c_1,...,c_k)$, and for input $X\in(\R^d)^n$, let $X_i\subseteq X$ be the cluster of center $c_i$ i.e.\ the points from the input $X$ that are closest to center $c_i$. The cost function for the $k$-center problem is defined by
$$\cost(C)=\max_{i\in [k]}{\left\{ \max_{x\in X_i}{\|x-c_i\|} \right\} }$$
\end{definition}
Note that this problem is fundamentally different from $k$-means and $k$-median since its cost function is much more sensitive. Where $k$-centers problem cost may change by $O(\Lambda)$ upon a change of a single point, $k$-means and $k$-median may change by only $O(\Lambda^2/n)$, $ O(\Lambda/n)$ respectively. 
For $k$-centers problem, Gonzalez (1985, \cite{gonzalez1985clustering}) showed an efficient 2-approximation algorithm.
In 1988, Feder and Greene \cite{feder1988optimal} showed that it is NP-hard to approximate the $k$-centers problem to a factor of $\alpha<1.822$.
\newpage

%%%%%%%%%%%%%%%
% Preliminaries
%%%%%%%%%%%%%%%
\section{Preliminaries}
\label{sec:preliminaries}
We require the following two folklore lemmas. The first lemma quantifies the $1$-means cost of a center $\hat{c}$ in terms of its distance from the optimal center.

\begin{lemma}
\label{lem:Ost22}
Let $X\in(\mathbb{R}^d)^n$, let $c$ denote the average of $X$. For any $\hat{c}\in\mathbb{R}^d$ it holds that
$$
\sum_{x\in X}\|x-\hat{c}\|^2=n\cdot\|\hat{c}-c\|^2+\sum_{x\in X}\|x-c\|^2.
$$
\end{lemma}

\begin{proof} (Kanungo et al.\ 2004, Lemma 2.1 \cite{kanungo2004local}, or Awasthi 2013, Fact 2.3.1 \cite{awasthi2013approximation})
\begin{align*}
\sum_{x\in X}{\|x-\hat{c} \|^2} 
&= \sum_{x\in X}{\sum_{i\in d}{|x_i - \hat{c}_i}|^2} \\
&= \sum_{x\in X}{\sum_{i\in d}{\left( |x_i - c_i|^2 + | c_i - \hat{c}_i |^2 + 2(x_i - c_i)\cdot(c_i - \hat{c}_i) \right)}}  \\
&= \sum_{x\in X}{\|x-c\|^2} + |X|\|c-\hat{c}\|^2 + \sum_{i\in d}{2(c_i-\hat{c}_i)\sum_{x\in X}{(x_i - c_i)}} \\
&= \sum_{x\in X}{\|x-c\|^2} + |X|\|c-\hat{c}\|^2
\end{align*}
where the last equality holds since $\forall i\in [d]. \sum_{x\in X} {(x_i - c_i)} = 0$
\end{proof}

The following is immediate corollary
\begin{corollary}
\label{cor:Ost22}
Let $X\in(\mathbb{R}^d)^n$, let $c$ denote the average of $X$. the following holds: $\sum_{\{x_1, x_2\}\subset X}{\|x_1 - x_2\|^2} = 	|X|\sum_{x\in X}{\|x - c\|^2}$
\end{corollary}

\begin{proof}
\begin{align*}
\sum_{\{x_1, x_2\} \subset X}{\|x_1 - x_2\|^2} 
&=\frac{1}{2}\sum_{\langle x_1, x_2 \rangle \in X \times X}{\|x_1 - x_2\|^2}\\
&=\frac{1}{2}\sum_{x_1 \in X}{ \sum_{x_2\in X}{\|x_1 - x_2\|^2}} \\
&=\frac{1}{2}\sum_{x_1 \in X}{ \left( |X|\cdot \|c - x_1\|^2 + \sum_{x_2\in X}{\|c - x_2\|^2} \right) } \\
&= \frac{1}{2}\left( |X|\cdot \sum_{x_1 \in X}{\|c - x_1\|^2} + \sum_{x_1 \in X}{\sum_{x_2\in X}{ \|c - x_2\|^2}} \right) \\
&= |X|\cdot \sum_{x\in X}{\|x-c\|^2}
\end{align*}
\end{proof}

The next lemma bounds the distance from the average of $X$ to the average of a subset of $X$.

\begin{lemma}[Lemma 2.3 \cite{DBLP:journals/jacm/OstrovskyRSS12}]
\label{lem:Ost23}
Let $X$ be a finite set of points in $\mathbb{R}^d$ and let $S\subseteq X$ with $S\neq\emptyset$. Let $c$ and $s$ denote the means of $X$ and $S$, resp. Then,
$$
\|s-c\|^2\leq\frac{\opt^2_1(X)}{|X|}\cdot\frac{|X\setminus S|}{|S|}.
$$
\end{lemma}

\begin{proof}
We first fix an arbitrary partition of $X$, $X_1\uplus X_2 = X$ ($X_1, X_2\neq \emptyset$). Denote $c_1, c_2$ as the corresponding means. Next we establish the following equation: 
\begin{equation}
  \label{eq:lem:Ost23_eq0}
  |X_1|\|c-c_1\|^2 + |X_2|\|c-c_2\|^2 = \frac{|X_1||X_2|}{|X|}\cdot \|c_1 - c_2\|^2
\end{equation}

Equation~\eqref{eq:lem:Ost23_eq0} holds by observing the following: define vector set $\widetilde{X}$ consisting of $|X_1|$ vectors located at $c_1$ and $|X_2|$ vectors located at $c_2$, and observe that mean of $\widetilde{X}$ is $c$ (the mean of $X$). For $\widetilde{X}$ the following holds: $|X_1|\|c-c_1\|^2 + |X_2|\|c-c_2\|^2 = \opt_1^2(\widetilde{X}) = \frac{1}{|\widetilde{X}|}\cdot \sum_{\{x_1,x_2\}\subset \widetilde{X}}{\|x_1 - x_2\|^2} = \frac{|X_1||X_2|}{|X|}\cdot \|c_1 - c_2\|^2$, 
%\begin{align*}
%|X_1|\|c-c_1\|^2 + |X_2|\|c-c_2\|^2
%&= \opt_1^2(\widetilde{X}) \\
%&= \frac{1}{|X|}\cdot \sum_{\{x_1,x_2\}\subset X}{\|x_1 - x_2\|^2}\\
%&= \frac{|X_1||X_2|}{|X|}\cdot \|c_1 - c_2\|^2
%\end{align*}
Where the first equality is due to $\widetilde{X}$ definition and the fact that the mean of cluster points is the optimum center for the cluster, and the second equality is due to Corollary~\ref{cor:Ost22}, thus establishing Equation~\eqref{eq:lem:Ost23_eq0}.\\

For $X$ the following holds:
\begin{align*}
\opt_1^2(X)
&= \sum_{x\in X}{\|x-c\|^2}\\
&= \sum_{x\in X_1}{\|x-c\|^2} + \sum_{x\in X_2}{\|x-c\|^2}\\
&\stackrel{(1)}{=} |X_1|\|c-c_1\|^2+\opt_1^2(X_1) + |X_2|\|c-c_2\|^2 + \opt_1^2(X_2)\\
&\stackrel{(2)}{=} \opt_1^2(X_1) + \opt_1^2(X_2) + \frac{|X_1||X_2|}{|X|}\cdot \|c_1 - c_2\|^2\\
&\geq \frac{|X_1||X_2|}{|X|}\cdot \|c_1 - c_2\|^2
\end{align*}
where Equality (1) is from Lemma~\ref{lem:Ost22}, and Equality (2) is by plugging in Equation~\eqref{eq:lem:Ost23_eq0}. And hence:
\begin{align}
\|c_1 - c_2\|^2 \leq \opt_1^2(X)\cdot \frac{|X|}{|X_1||X_2|}\label{eq:lem:Ost23_eq1}
\end{align}
Observing $|X|c = |X_1|c_1 + |X_2|c_2$ yields $c_2 = \frac{|X|}{|X_2|} c - \frac{|X_1|}{|X_2|} c_1$ which we plug into Equation~\eqref{eq:lem:Ost23_eq1} left hand side to get:
\begin{align*}
\|c_1 - c_2\|^2 
&= \left\| c_1 - \left( \frac{|X|}{|X_2|} c - \frac{|X_1|}{|X_2|} c_1\right) \right\|^2\\
&= \left\| \frac{|X_2|+|X_1|}{|X_2|}c_1 -  \frac{|X|}{|X_2|} c \right\|^2\\
&= \left( \frac{|X|}{|X_2|} \right)^2 \| c_1 - c \|^2\\
\end{align*}
Reorganizing completes the proof:
$$\|c_1 - c\|^2 \leq \frac{\opt_1^2(X)}{|X|}\frac{|X_2|}{|X_1|} = \frac{\opt_1^2(X)}{|X|}\frac{|X\setminus X_1|}{|X_1|}$$

\end{proof}

We now show an argument that is ubiquitously used in clustering literature. Given a set of points, it is useful to bound the number of them that are far from their mean by some radius, the radius is normally taken as a factor of their standard deviation. Since the bound is essentially Markov inequality, it is sometimes referred as `Markovian argument'. Formally:
\begin{Claim}\label{clm:markovArg}
Consider $S\in (\R^d)^n$ and its mean $\mu_{S} =\frac{1}{n} \sum_{x\in S}{x}$. Denote for any $x\in \R^d$ $d(x)=\|x-\mu_{S}\|_2$. Let $s\in_{R} S$ and denote $r$ as follows:
$$Var(s) = \frac{1}{|S|}\sum_{x\in S}{\|x-\mu_{S}\|^2} = \frac{1}{|S|}\sum_{x\in S}{d^2(x)} \triangleq r^{2}.$$
Therefore $r$ is $s$'s standard deviation. Denote the set of points from $S$ that are far from $\mu_{S}$ by some factor as $S^{far}(a)=\left\{x\in S| d(x)\geq \frac{r}{a} \right\}$ for some $a\in \R^+$, and $n^{far}(a)=|S^{far}(a)|$ then the following holds:
$$n^{far}(a) \leq a^2 \cdot n$$
\end{Claim}
%\doteq
\begin{proof}
for some $a\in \R^+$
\begin{align*}
\frac{n^{far}(a)}{n} &= \Pr_{s\in_R S}{\left[  s\in S^{far}(a)\right]} = \Pr_{s\in_R S}{\left[  d
(s) \geq \frac{r}{a}\right]}\\
&= \Pr_{s\in_R S}{\left[  d^2
(s) \geq \left (\frac{r}{a} \right)^2 \right]} = \Pr_{s\in_R S}{\left[  d^2
(s) \geq \frac{\mathbb{E} \left[ d^2(s)\right]}{a^2} \right]} \leq a^2
\end{align*}
where the inequality is due to Markov inequality.
\end{proof}

\begin{Claim}[Sum of squares bound]\label{clm:sosBound}
Let $x,y\in \R^d$ then the following holds:
$$\| x+y \|^2 \leq 2\cdot \left( \|x\|^2 + \|y\|^2 \right) $$  
\end{Claim}

\begin{proof} Let $x,y\in \R^d$. The following holds:
\begin{align*}
\|x+y\|^2 &\leq (\|x\|+\|y\|)^2 \\
&=\|x\|^2 + \|y\|^2 + 2\|x\|\|y\| \\
&\leq \|x\|^2 + \|y\|^2 + \|x\|^2 + \|y\|^2 \\
&= 2(\|x\|^2 + \|y\|^2)
\end{align*}
Where the second inequality is due to the fact that for any $a,b\in \R$ it holds that $2ab \leq a^2 + b^2$. To see that, observe that for any $a,b\in \R$ it holds that $a^2-2ab+b^2=(a-b)^2 \geq 0$.
\end{proof}

%\begin{proof} We look on the plane spanned by $x,y$, and denote $z=x+y$. then $z\in \vecspan\{ x,y \}$. Also denote by superscript $t,p$ the perpendicular and tangential components of vectors w.r.t the line formed by $z$. Then for any $x,y \in \R^d$ it holds that $\| x^t \| + \| y^t \| = \| z \|$. We first bound the following:
%$$\|x^t\|^2 + \| y^t \|^2 = \|x^t\|^2 + \left(\|z\| - \|x^t\| \right)^2 = 2\cdot \|x^t\|^2 - 2\cdot \|z\|\|x^t\| + \|z\|^2$$
%We have a quadratic formula w.r.t $\|x^t\|$, that have a single optimum (minimum) achieved on the value $\|x^t\| = \frac{1}{2}\|z\|$. Therefore $\|x^t\|^2 + \| y^t \|^2 \geq \frac{1}{2} \| z \|^2$. To complete the proof we use Pythagoras theorem:
%\begin{align*}
%\|x\|^2 + \|y\|^2 &= \|x^p\|^2 + \|x^t\|^2 + \|y^p\|^2 + \|y^t\|^2\\
%&\geq \|x^t\|^2 + \|y^t\|^2 \\
%&\geq \frac{1}{2} \| z \|^2 \\
%&= \frac{1}{2} \| x+y \|^2
%\end{align*}
%\end{proof}

\subsection{Clustering under stability assumptions}

%In recent years, many works have studied the notion of clustering under various input-stability assumptions~\cite{DBLP:journals/jacm/OstrovskyRSS12, AckermanB09, BalcanBG09, BiluL10, AwasthiBS10, KumarK10, AwasthiBS12}, showing how to rely on input-niceness in order to obtain a good approximation and even a PTAS~\cite{Cohen-AddadS17, AwasthiBS10}, for clustering problems. The main focus of this work is the stability notion of Ostrovsky et al.~\cite{DBLP:journals/jacm/OstrovskyRSS12} who defined a clustering instance to be {\em well-separated} for $k$ clusters if the optimal partitioning of the data into $k$ clusters has cost noticeably smaller than the cost of any partitioning of the data into $k-1$ clusters (see Definition~\ref{def:seperationCriteria}).\\

The following theorem relates the notion of approximating the $k$-means cost and approximating the true $k$-means centers in Wasserstein distance for $\phi$-separated instances. For completeness we include its proof.

\begin{theorem}[{{\cite[Theorem 5.1]{DBLP:journals/jacm/OstrovskyRSS12}}}]\label{thm:ostKmeans}
Let $\alpha$ and $\phi$ be such that $\frac{\alpha+\phi^2}{1-\phi^2}<\frac{1}{16}$. Suppose that $X\subseteq\R^d$ is $\phi$-separated for $k$-means, let $C=(c_1,\dots,c_k)$ be a set of optimal centers for $X$, and let $\hat{C}=(\hat{c}_1\dots,\hat{c}_k)$ be centers such that $\cost_X(\hat{C})\leq\alpha\cdot\opt_{k-1}(X)$. Then for each $\hat{c}_i$ there is a distinct optimal center, call it $c_i$, such that $\|\hat{c}_i-c_i\|\leq 2\cdot\sqrt{\frac{\alpha+\phi^2}{1-\phi^2}} \cdot D_i$, where $D_i = \min_{j\neq i}\|c_i-c_j\|$. 
\end{theorem}

\begin{proof}
Let $\rho=(\frac{\alpha}{\phi^2}+1)^{-1}$. For each $i\in[k]$ define $r_i=\sqrt{\frac{1}{n_i}\sum_{x\in X_i}\|x-c_i\|^2}$, and $X_i^{\rm cor}=\{x\in X_i: \|x-c_i\|\leq\frac{r_i}{\sqrt{\rho}}\}$. 
A standard argument (see Claim~\ref{clm:markovArg}) shows that $|X_i^{\rm cor}|\geq(1-{\rho})n_i$. Let $d^2_i=\phi^2\opt_{k-1}(X)/n_i$. We argue that $r_i^2\leq d_i^2\leq\frac{\phi^2}{1-\phi^2}\cdot D_i^2$. Indeed,
$$
r_i^2=\frac{1}{n_i}\sum_{x\in X_i}\|x-c_i\|^2\leq\frac{1}{n_i}\cdot\opt_k(X)\leq\frac{1}{n_i}\phi^2\opt_{k-1}(X)=d_i^2,
$$
which shows the first inequality. To see that $d_i^2\leq\frac{\phi^2}{1-\phi^2}\cdot D_i^2$, we first show the following:
$$
\opt_{k-1}(X)\leq\opt_k(X)+n_i\cdot D_i^2
$$
Above holds since by Lemma~\ref{lem:Ost22} the right term is equal to a cost of the following $k-1$  centers: $C\setminus\{c_i\}$, with the following assignment: each cluster $j\in [k]\setminus\{i\}$ is assigned to its center, and the points of cluster $i$ are assigned to the closest center of $c_i$. That  assignment is a $k-1$ cost, that is lower bounded by the optimal $k-1$ cost, the right term, thus establishing the inequality.\\

Now, by the input assumption we have $\opt_k(X)+n_i\cdot D_i^2 \leq \phi^2\opt_{k-1}(X)+n_i\cdot D_i^2$. These two inequalities yields $\opt_{k-1}(X)\leq n_i\cdot D_i^2/(1-\phi^2)$, and hence, 
$$
d_i^2=\phi^2\opt_{k-1}(X)/n_i\leq \frac{\phi^2}{1-\phi^2}\cdot D_i^2.
$$

We say that a center $\hat{c}_i$ is {\em close} to an optimal center $c_j$ if $\|\hat{c}_i-c_j\|\leq2\sqrt{\frac{\alpha+\phi^2}{1-\phi^2}}\cdot D_j$. Observe that if $2\sqrt{\frac{\alpha+\phi^2}{1-\phi^2}}<1/2$ then a center $\hat{c}_i$ can be close to at most one optimal center. 
Assume towards contradiction that there is a center $\hat{c}_i$ such that $\hat{c}_i$ is not close to any optimal center. Therefore, by the pigeonhole principle, there must exist an optimal center $c_j$ that is not close to any center in $\hat{C}$.
Then, in the clustering around $\hat{c}_1,\dots,\hat{c}_k$, all the points in $X_j^{\rm cor}$ are assigned to a center that is more than $2\sqrt{\frac{\alpha+\phi^2}{1-\phi^2}}\cdot D_j$ away from $c_j$. Recall that $X_j^{\rm cor}$ contains all points whose distance to $c_j$ is at most 
$$\frac{r_j}{\sqrt{\rho}}\leq \frac{\sqrt{\frac{\phi^2}{1-\phi^2}\cdot D^2_j}}{\sqrt{\rho}}=\sqrt{\frac{\alpha+\phi^2}{1-\phi^2}}\cdot D_j.$$
Hence, in the clustering around $\hat{c}_1,\dots,\hat{c}_k$, all the points in $X_j^{\rm cor}$ are assigned to a center that is more than $\sqrt{\frac{\alpha+\phi^2}{1-\phi^2}}\cdot D_j$ away from them. Therefore,
\begin{align*}
\cost_X(\hat{C})
&> |X_j^{\rm cor}|\cdot \frac{\alpha+\phi^2}{1-\phi^2}\cdot D^2_j\\
&\geq \left(1-\rho\right)n_j \cdot \frac{\alpha+\phi^2}{1-\phi^2}\cdot D^2_j\\
&= \alpha \cdot n_j \cdot \frac{D^2_j}{1-\phi^2}\\
&\geq \alpha \cdot n_j \cdot \frac{d^2_j}{\phi^2}\\
&= \alpha\cdot\opt_{k-1}(X),
\end{align*}
giving a contradiction.
\end{proof}

Next we show equivalence between several different notions of stability:

\begin{lemma}
\label{lem:StabilityNotionsEquivalence}
Given a $k$-clustering objective in the form $\min\limits_{C} \sum_{x\in X}\min_{c\in C} \|x- c\|^p$ for $p\in\{1,2\}$, the following notions of stability are all equivalent up to a constant factor.
\begin{enumerate}
	\item[{\bf 1.}] {\bf $\boldsymbol\phi$-well separability~\cite{DBLP:journals/jacm/OstrovskyRSS12}:} $\opt_k \leq \phi^p\cdot\opt_{k-1}$.
	\item[{\bf 2.}] {\bf $\boldsymbol\beta$-center deletion~\cite{AwasthiBS10}:} For every cluster $i$ and $j\neq i$, delete center $c_i$ and assign all of its points to center $c_j$. The result is a $(k-1)$-clustering of cost $\geq \beta \opt_k$.
	\item[{\bf 3.}] {\bf $\boldsymbol\gamma$-center separation~\cite{AwasthiBS10}:} For every cluster $i$, denote its size by $|X_i|$ and let $D_i^p = \min_{j\neq i} \|c_i-c_j\|^p$. Then $D_i^p \geq \frac{\gamma}{|X_i|} \opt_k$.
	\item[{\bf 4.}] {\bf $\boldsymbol(\delta,\tfrac 1 4)$-approximation-center stability~\cite{DBLP:journals/jacm/OstrovskyRSS12}:} For any $k$-tuple $\hat c_1,.., \hat c_k$ of cost at most $\delta \opt_k$, we have a matching $\varphi$ such that $\|c_i-\hat{c}_{\varphi(i)}\|^p < \tfrac 1 4\cdot D_i^p$.\footnote{We comment that we can replace the constant $\tfrac 1 4$ with any constant $<\tfrac 1 2$.}
\end{enumerate}
\end{lemma}
\begin{proof}
	\underline{$1\Rightarrow 2$:} By assigning all points in $X_i$ to  center $c_j$ we get some $k-1$ clustering, with cost $\geq \opt_{k-1}\geq \phi^{-p}\opt_k$. Thus the input is $\phi^{-p}$-center deletion.
	
\medskip
\noindent\underline{$2\Rightarrow 3$:} Fix $i$. By assigning all points in $X_i$ to center $c_j$ the cost of this $k-1$ clustering is at most $\opt_k + |X_i| D_i^p$.\footnote{Here is where the analysis slightly deviates for $p=1$, where we just use triangle inequality, and for $p=2$ where we use the properties of the $k$-means cost when shifting a cluster's mean i.e.\ Lemma~\ref{lem:Ost22}} It follows that $|X_i|D_i^p \geq (\beta-1)\opt_k$, implying we have a $(\beta-1)$-center separation.
	
\medskip
\noindent\underline{$3\Rightarrow 4$:} We argue the contrapositive. Fix $\hat c_1,.., \hat c_k$ of cost $\delta\opt_k$ for $\delta \leq \frac \gamma 8 - 1$ such that for some $i$ we have that $\min_j \|c_i-\hat c_j\|^p > \tfrac 1 4 D^p_i$. For each $x\in X_i$ we denote $\hat c(x)$ as the center it is assigned to, and we have that the contribution of the points in $X_i$ is
	\begin{align*}
	\sum_{x\in X_i} \|x-\hat c(x)\|^p &=  \sum_{x\in X_i} \| \hat c(x)-c_i - (x-c_i)\|^p \stackrel{(1)}\geq \sum_{x\in X_i} \tfrac 1 p \|\hat c(x)-c_i\|^p - \|x-c_i\|^p  \cr
	&\geq \tfrac {|X_i|} {4p} D_i^p - |X_i| r_i^p\geq \frac{\gamma} {4p} \opt_k -\opt_k = (\frac {\gamma}{4p} -1)\opt_k\geq\delta\opt_k
	\end{align*}
	yielding a contradiction. The inequality marked by $(1)$ follows from the standard triangle inequality for $p=1$ and the fact that $(a-b)^2 \geq \tfrac 1 2 a^2 -b^2$ for $p=2$.
	Thus the input is ($\tfrac \gamma {4p} - 1, \tfrac 1 4$)-approximation stable.
	
\medskip
\noindent\underline{$4\Rightarrow 1$:} Assume for contradiction there exists some $(k-1)$-tuple of centers $(\hat c_1,...,\hat c_{k-1})$ with cost $<\delta \opt_k$. Create a $k$-tuple of centers by adding a point arbitrarily far from all other $k-1$ centers. We obtain a $k$-clustering of cost $< \delta \opt_k$ implying some optimal center $c$ must be matched with the arbitrarily far point we added, contradicting the approximation stability. It follows this is a $\delta^{-1/p}$-well separated instance.  
\end{proof}

As an immediate corollary, it follows that our algorithms are applicable to any instance satisfying one of the above mentioned stability notions (with suitable stability parameters).

\subsection{Preliminaries from differential privacy}
The following theorems, and other fundamental results in differential privacy field are presented in the book `The Algorithmic Foundations of Differential Privacy' by Dwork and Roth~\cite{DR14}. The most basic constructions of differentially private algorithms are via the Laplace and Gaussian mechanisms as specified in the following theorems.

\begin{definition}[$L_p$-Sensitivity]
A function $f$ mapping databases to $\R^d$ has {\em $L_p$-sensitivity $\lambda$} if $\|f(S)-f(S')\|_p\leq \lambda$ for all neighboring $S,S'$.
\end{definition}

\begin{theorem}[Laplace mechanism~\cite{dwork2006calibrating}]\label{thm:lap}
A random variable is distributed as $\Lap(b)$ if its probability density function is $h(y)=\frac{1}{2b}\exp(-\frac{|y|}{b})$.
Let $\eps>0$, and let $f:U^n \rightarrow \R^d$ be a function of $L_1$-sensitivity $\lambda$. The mechanism $\AAA$ that on input $S\in U^n$ outputs $f(S) + \left(\Lap(\frac{\lambda}{\eps})\right)^d$ is $(\eps,0)$-differentially private.\footnote{For a distribution $\DDD$ we write $\DDD^d$ to denote the product distribution defined by sampling $(x_1,...,x_d)$ where each $x_i$ is sampled independently from $\DDD$.} 
%Moreover, $\Pr\Big[ \|\AAA(S)-f(S)\|_{\infty}>\Delta \Big]\leq d\cdot\exp\left(-\frac{\eps \Delta}{\lambda}\right)$.
\end{theorem}

\begin{theorem}[Gaussian Mechanism \cite{DKMMN06}]\label{thm:gauss}
Let $\eps,\delta\in(0,1)$, and assume $f:U^n \rightarrow \R^d$ has $L_2$-sensitivity $\lambda$. Let $\sigma\geq\frac{\lambda}{\eps}\sqrt{2\ln(1.25/\delta)}$.
The mechanism that on input $S\in U^n$ outputs $f(S)+\left(\NNN(0,\sigma^2)\right)^d$ is $(\eps,\delta)$-differentially private.
\end{theorem}

In order to reason about the privacy loss of a {\em single} individual that is a member of a database we use the notion of differential privacy (Definition~\ref{def:dp_intro}), which bounds that loss by the privacy parameters $(\eps, \delta)$. It also makes sense to reason about privacy loss of a {\em group of individuals}, in order to bound their privacy loss in the same sense. The following theorem gives such bounds: 
\begin{theorem}[Group Privacy \cite{Vadhan2016}]\label{thm:groupPrivacy}
Let $\XXX$ be a domain of record type and let $\AAA : \N^{\XXX} \rightarrow Y$. If $\AAA$ is an $(\eps,\delta)$-{\em differentially private} mechanism, then for any pair of databases $X,X'\in \N^{\XXX}$ that differ by at most $k$ rows (i.e. $\|X-X'\|\leq k, \|X\|=\|X'\|$), and every set $T\subseteq Y$, we have 
$$\Pr[\AAA(X)\in T]\leq e^{k\eps}\cdot \Pr[\AAA(X')\in T]+k\cdot e^{k\eps}\cdot\delta.$$
\end{theorem}

Mechanisms (randomized algorithms) that admit differential privacy enjoy a very useful feature - running them in a parallel or a sequential manner results in another (composed) differentially private mechanism. That feature, known in the literature as {\em Composition theorems}, quantifies the composed mechanism's promised privacy parameter $(\eps, \delta)$. A desirable property for the composition is {\em adaptivity}. Specifically, we allow the choice of the next (differentially private) mechanism to be applied to depend on the outcomes of the previous (differentially private) mechanisms we applied. The following theorems quantify such adaptive composition.

%\begin{theorem}[Simple composition~\cite{DKMMN06, DBLP:conf/stoc/DworkL09}]\label{thm:compSimple} Let $M_i:\XXX^n \xrightarrow{}S_i$ be an $(\eps_i,\delta_i)$-differentially private algorithm for $i\in [k]$. Then if $M:\XXX^n \xrightarrow{} (S_1 \times ... \times S_k)$ is defined to be $M(x)=\langle M_1(x), ..., M_k(x)\rangle$, then $M(x)$ is $(\sum_{i\in [k]}{\eps_i},\sum_{i\in [k]}{\delta_i})$-differentially private.
%\end{theorem}

%Above theorem allows quantification of the privacy parameters of a mechanism that is composed by other $k$ private mechanism, but their runs are independent from one another. We would also like to quantify privacy parameters of a mechanism that chooses its composing mechanisms, their input databases and each of the runs parameters in an {\em adaptive} manner. That is, that the built mechanism $M$ can choose its next $i$'th mechanism $M_i$, its data base and the run parameters w.r.t all of the information revealed to it up until its current stage. The following theorems quantifies such adaptive composition:

\begin{theorem}[Simple composition~\cite{DKMMN06, DBLP:conf/stoc/DworkL09}]\label{thm:compSimple} Let $M:\XXX^n \xrightarrow{} (S_1 \times ... \times S_k)$ be a mechanism that permits $k$ adaptive interactions with $(\eps, \delta)$-differentially private mechanisms (and does not access the data base otherwise) then $M$ is $(k\eps, k\delta)$-differentially private.
\end{theorem}

\begin{theorem}[Advanced composition \cite{dwork2010boosting}]\label{thm:compAdvanced} Let $M:\XXX^n \xrightarrow{} (S_1 \times ... \times S_k)$ be a mechanism that permits $k$ adaptive interactions with $(\eps, \delta)$-differentially private mechanisms (and does not access the data base otherwise), then for any $\eps,\delta'\in (0,1]$, $\delta\in [0,1]$,  $M(x)$ is $(\eps', k\delta + \delta')$-differentially private where $\eps'=\sqrt{2k\ln{(1/\delta')}}\eps + k\eps(e^{\eps}-1)$
\end{theorem}

Theorem~\ref{thm:compSimple} states that a  composition of $k$ mechanisms, each preserves $(\eps,\delta)$-differential privacy, results in a mechanism that preserves $(k \eps, k \delta)$-differential privacy. That is, the privacy guarantees degrade at most linearly with the number of composed mechanisms.
Theorem~\ref{thm:compAdvanced}  presents a different bound on the resulting privacy guarantees. Informally, it states that the privacy guarantees degrade only proportionally to the square root of the number of composed mechanisms.
Theorems \ref{thm:compSimple}, \ref{thm:compAdvanced} are valid simultaneously.\\

The following theorem states that running an $(\eps, \delta)$-differentially private algorithm on only a sub-sample of its database, results in a private algorithm:

\begin{theorem}[Sampling with replacement \cite{bun2015differentially}]\label{thm:sampWithReplacment} Fix $\eps\leq 1$ and let $\AAA$ be an $(\eps,\delta)$-differentially private algorithm operating on databases of size $m$. For $n\geq 2m$, construct an algorithm $\tilde{\AAA}$ that on an input of a database $D$ of size $n$ subsamples (with replacement) $m$ rows from $D$ and runs $\AAA$ on the result. Then $\tilde{\AAA}$ is $(\tilde{\eps}, \tilde{\delta})$-differentially private for:
$$\tilde{\eps}=6\eps m/n \text{  and  } \tilde{\delta}=\exp(6\eps m/n)\frac{4m}{n}\cdot \delta.$$
\end{theorem}

% Below is taken from arXiv version v10.5

%%%%%%%%%%%%%%%%%%%%%%%%%%%%%%%%%%%%%%
% $k$-means with stability assumption
%%%%%%%%%%%%%%%%%%%%%%%%%%%%%%%%%%%%%%
\newpage
\section{Stability Improves Accuracy for Private Clustering Algorithms}
\label{sec:main}

In this section we show that applying a private clustering algorithm with a worst-case guarantee of $v$-approximation on a $\phi$-well separable instance results in (effectively) a $(1+O(\phi^2))$-approximation for the $k$-means, provided that $v$ is small in comparison to $\phi^{-2}$. In other words, we show that when running a private clustering algorithm $\AAA$ on {\em stable instances}, then $\AAA$ actually performs much better than its worst case bounds. We focus here on the $k$-means cost objective, and present an analogues result for $k$-median in Section~\ref{sec:median}. Our construction appears in Algorithm \texttt{Private-Stable-$k$-Means}.

\begin{algorithm*}[htbp]

\caption{\texttt{Private-Stable-$k$-Means}}\label{alg:privateKmeans}

{\bf Input:} Database $X$ containing $n$ points in the $d$-dimensional ball $\BBB(0,\Lambda)$, failure probability $\beta$, privacy parameters $\eps,\delta$.

{\bf Tool used:} An $(\eps,\delta)$-differentially private algorithm $\AAA$ for approximating the $k$-means.

\begin{enumerate}[leftmargin=15pt,rightmargin=10pt,itemsep=1pt,topsep=0pt]

\item Run $\AAA$ on $X$ to obtain $k$ centers: $B=\{b_1,\dots,b_k\}$.

\item For $i\in[k]$ let $\hat{D}_i=\min_{j\neq i}\|b_i-b_j\|$.

\item For $i\in[k]$ let $\hat{X}_i=\{x\in X : \|x-b_i\|\leq\hat{D}_i/3\}$.

\item Let $\overline{C}=\{\overline{c}_1,\dots,\overline{c}_k\}$ denote the average of the points in $\hat{X}_1,\dots,\hat{X}_k$, respectively. For $i\in[k]$ use the Gaussian mechanism (see \cite{DKMMN06}) with privacy parameters $(\eps,\delta)$ to compute a noisy estimation $\hat{c}_i$ of $\overline{c}_i$. Denote $\hat{C}=\{\hat{c}_1,\dots,\hat{c}_k\}$.

\item Use the Gaussian mechanism with privacy parameters $(\eps,\delta)$ to estimate $\cost_X(\hat{C})$ and $\cost_X(B)$. Output the set of centers (either $\hat{C}$ or $B$) with the lower (estimated) cost.

\end{enumerate}
\end{algorithm*}

Algorithm \texttt{Private-Stable-$k$-Means} applies $4$ $(\eps,\delta)$-differentially private mechanisms and therefore it is $(4\eps, 4\delta)$-differentially private. That is immediate from composition properties of differential privacy (see Theorem~\ref{thm:compSimple}, \cite{DKMMN06, DBLP:conf/stoc/DworkL09}).
We proceed with its utility analysis. 
Let $X$ be $\phi$-separated for $k$-means with optimal centers $C^*=\{c^*_1,\dots,c^*_k\}$, and let $X^*_1,\dots,X^*_k\subseteq X$ be the clusters induced by $C^*$. For $i\in[k]$ we denote $n_i=|X^*_i|$ and $r^*_i=\sqrt{\frac{1}{n_i}\sum_{x\in X^*_i}\|x-c^*_i\|^2}$. Finally, denote $\rho=\frac{100\phi^2}{1-\phi^2}$.\footnote{We note that the analysis holds for a range of values of $\rho$, and the value $\rho=\frac{100\phi^2}{1-\phi^2}$ was chosen w.r.t constrains added in the analysis of Lemma \ref{lem:lem1}. More specifically, if $\rho=\frac{a\cdot\phi^2}{1-\phi^2}$, then the analysis holds for any $a>9$ and present the following trade-off between the resulting multiplicative factor of Theorem~\ref{thm:main} and the required parameter $\phi$: for small values of $a$ we get a better multiplicative factor for Theorem~\ref{thm:main} (the constant that is hiding in the `$O$' notation is linear with $a$), while imposing a harder requirement (smaller value) for $\frac{\phi^2(w+1)}{1-\phi^2}$, thus resulting with a requirement for a smaller $\phi$ for the input (and vice versa).}
Consider the execution of \texttt{Private-Stable-$k$-Means} on $X$, and let $B=\{b_1,\dots,b_k\}$ and $\overline{C}=\{\overline{c}_1,\dots,\overline{c}_k\}$ denote the centers obtained in Steps~1 and~4.   We assume for simplicity (and without loss of generality) that the set of optimal centers $C^*=\{c^*_1,\dots,c^*_k\}$ is sorted s.t.\ $c^*_i$ is the closest to $b_i$. We note that such a matching exists provided that the requirements of Theorem~\ref{thm:ostKmeans} are met. The next lemma shows that, provided that $B$ has a low enough cost, the distance from each $\overline{c}_i$ to its corresponding optimal center is low.

\begin{lemma}\label{lem:lem1}
If $\cost_X(B)\leq w\cdot\phi^2\cdot\opt_{k-1}(X)$ and if $\frac{\phi^2(w+1)}{1-\phi^2}$ is sufficiently small (specifically: $\frac{\phi^2(w+1)}{1-\phi^2}\leq \frac{1}{784}$), then
$$\|\overline{c}_i-c^*_i\|^2\leq {(r_i^*)}^2\cdot\frac{\rho}{1-\rho}$$
%$$\|\overline{c}_i-c^*_i\|^2\leq\frac{\opt_1(X^*_i)}{n_i}\cdot\frac{\rho}{1-\rho}$$
\end{lemma}

\begin{proof}
For $i\in[k]$ define $X_i^{\rm cor}=\{x\in X^*_i: \|x-c^*_i\|\leq\frac{{r^*_i}}{\sqrt{\rho}}\}$.
Standard Markovian argument (see Claim~\ref{clm:markovArg}) shows that $|X_i^{\rm cor}|\geq(1-\rho)\cdot n_i$. We first show that for every $i\in[k]$ we have $X_i^{\rm cor}\subseteq \hat{X}_i\subseteq X^*_i$. To that end, fix $i\in[k]$ and recall that $\cost_X(B)\leq w\cdot\phi^2\cdot\opt_{k-1}(X)$.
Denote $\gamma=2\sqrt{\frac{\phi^2(w+1)}{1-\phi^2}}$, and $D_i=\min_{j\neq i}\|c^*_i-c^*_j\|$. By Theorem~\ref{thm:ostKmeans} we have that 
$\|c^*_i-b_i\|\leq\gamma\cdot D_i$.

Now, $\hat{X}_i$ contains every point $x\in X$ whose within a distance from $b_i$ of 
\begin{align}
\frac{1}{3}\cdot\hat{D}_i
&=\frac{1}{3}\cdot\min_{j\neq i}\|b_i-b_j\|\nonumber\\
&\geq \frac{1}{3}\cdot\min_{j\neq i}\left(\|c^*_i-c^*_j\|-\|b_i-c^*_i\|-\|b_j-c^*_j\|\right)\nonumber\\
&\geq \frac{1}{3}\cdot\min_{j\neq i}\left(\|c^*_i-c^*_j\|-\gamma\cdot D_i-\gamma\cdot D_j \right)\nonumber\\
&\geq \frac{1}{3}\cdot\min_{j\neq i}\left(\|c^*_i-c^*_j\|-\gamma\cdot \|c^*_i-c^*_j\|-\gamma\cdot\|c^*_{i}-c^*_j\|\right)\nonumber\\
&= \frac{1-2\gamma}{3}\cdot\min_{j\neq i}\left(\|c^*_i-c^*_j\|\right)\nonumber\\
&=\frac{1-2\gamma}{3}\cdot D_i\label{eq:lem1_eq0}
\end{align}
In particular, $\hat{X}_i$ contains every point $x\in X$ whose within a distance from $c^*_i$ of
$$
\frac{1-2\gamma}{3}\cdot D_i-\gamma\cdot D_i = \frac{1-5\gamma}{3}\cdot D_i\geq \frac{1-5\gamma}{3}\cdot \sqrt{\frac{1-\phi^2}{\phi^2}}\cdot r^*_i\geq r^*_i/\sqrt{\rho},
$$
where the first inequality is from Theorem~\ref{thm:ostKmeans} and the second inequality holds for sufficiently small $\phi$ and $w$, specifically, $\frac{\phi^2(w+1)}{1-\phi^2}\leq \left(\frac{7}{100}\right)^2 = 0.0049$. Therefore,  $X_i^{\rm cor}\subseteq \hat{X}_i$ (because $X_i^{\rm cor}$ contains points within distance $r^*_i/\sqrt{\rho}$ from $c^*_i$). Similar arguments show that $\hat{X}_i\subseteq X^*_i$. Specifically, let $j_i=\argmin_{j\neq i}\|c^*_i-c^*_j\|$. Now, $X^*_i$ contains every point $x\in X$ whose within a distance from $c^*_i$ of
\begin{align*}
\frac{1}{2}D_i&=\frac{1}{2}\min_{j\neq i}\|c^*_i-c^*_j\|=\frac{1}{2}\|c^*_i-c^*_{j_i}\|\\
&\geq\frac{1}{2}\left(\|b_i-b_{j_i}\|-\|b_i-c^*_i\|-\|b_{j_i}-c^*_{j_i}\|\right)\\
&\geq\frac{1}{2}\left(\hat{D}_i-\gamma D_i-\gamma D_{j_i}\right)\\
&=\frac{1}{2}\left(\hat{D}_i-2\gamma D_i\right)\\
&\geq\frac{1}{2}\left(\hat{D}_i-\frac{2\gamma}{1-2\gamma} \hat{D}_i\right)
=\frac{1-4\gamma}{2-4\gamma}\cdot\hat{D}_i
\end{align*}
where the last inequality is from inequality~(\ref{eq:lem1_eq0}).
In particular, $X^*_i$ contains every point $x\in X$ whose within a distance from $b_i$ of 
$$
\frac{1-4\gamma}{2-4\gamma}\cdot\hat{D}_i - \|c^*_i-b_i\| \geq \frac{1-4\gamma}{2-4\gamma}\cdot\hat{D}_i - \gamma\cdot D_i \geq \frac{1-4\gamma}{2-4\gamma}\cdot\hat{D}_i - \frac{\gamma}{1-2\gamma}\cdot \hat{D}_i\geq\frac{1}{3}\cdot\hat{D}_i,
$$
where the last inequality holds for sufficiently small $\gamma$, specifically for $\gamma \leq \frac{1}{14}$, which is true whenever $\frac{\phi^2(1+w)}{1-\phi^2}\leq \frac{1}{784} \approx 0.0013$. Therefore, $\hat{X}_i\subseteq X^*_i$ (because $\hat{X}_i$ contains points within distance $\hat{D}_i/3$ from $b_i$). So,
$$
X_i^{\rm cor}\subseteq \hat{X}_i\subseteq X^*_i.
$$
Recall that $\overline{c}_i$ denotes the average of the points in $\hat{X}_i$. By Lemma~\ref{lem:Ost23} we have that
\begin{align*}
\|\overline{c}_i-c^*_i\|^2\leq\frac{\opt_1(X^*_i)}{|X^*_i|}\cdot\frac{|X^*_i\setminus \hat{X}_i|}{|\hat{X}_i|}\leq\frac{\opt_1(X^*_i)}{n_i}\cdot\frac{|X^*_i\setminus X_i^{\rm cor}|}{|X_i^{\rm cor}|}\leq\frac{\opt_1(X^*_i)}{n_i}\cdot\frac{\rho}{1-\rho}
\end{align*}
\end{proof}

Let $\hat{C}$ be the centers obtained in Step~4 of the execution, and recall that each $\hat{c}_i\in\hat{C}$ is a noisy estimation of $\overline{c}_i$, where $\overline{c}_i$ is the average of the points in $\hat{X}_i$ (all the input points whose distance to $b_i$ is significantly smaller than their distance to any other $b_j$). The next lemma shows that the $k$-means cost of $\hat{C}$ is low. This is done by relating the cost of $\hat{C}$ to that of $\overline{C}$, which we then relate to the cost of the optimal centers using Lemma~\ref{lem:lem1}. 

\begin{lemma}\label{lem:goodC}
If $\cost_X(B)\leq w\cdot\phi^2\cdot\opt_{k-1}(X)$ and if $\frac{\phi^2(w+1)}{1-\phi^2}$ is sufficiently small (specifically: $\frac{\phi^2(w+1)}{1-\phi^2}\leq \frac{1}{784}$), then
$$\cost_X(\hat{C})\leq \left(1+O(\phi^2)\right)\cdot\opt_k(X) + O(1)\cdot  \frac{k \Lambda^2 \sqrt{d}}{\eps}\cdot\ln\left(\frac{dk}{\beta\delta}\right).$$
\end{lemma}

\begin{proof}
First let us assume that for every $i\in[k]$ it holds that $|\hat{X}_i|\geq\frac{16}{\eps}\log(\frac{4k}{\beta\delta})$, which is true if $n_i=|X^*_i|\geq\frac{16}{\eps(1-\rho)}\log(\frac{4k}{\beta\delta})$. Fix $i\in[k]$. By the properties of the Gaussian mechanism (see \cite{DKMMN06}), with probability at least $(1-\frac{\beta}{k})$ we have that 
$$
\|\hat{c}_i-\overline{c}_i\|\leq\frac{64\Lambda\sqrt{d}}{\eps\cdot|\hat{X}_i|}\cdot\ln\left(\frac{8dk}{\beta\delta}\right)\leq\frac{64\Lambda\sqrt{d}}{\eps\cdot (1-\rho)n_i}\cdot\ln\left(\frac{8dk}{\beta\delta}\right).
$$
Thus, by Lemma~\ref{lem:Ost22} we have that
\begin{align*}
\cost_X(\hat{C})&\leq\sum_i\sum_{x\in X^*_i}\|x-\hat{c}_i\|^2\\
&=\sum_i\left(\opt_1(X^*_i)+n_i\cdot\|\hat{c}_i-c^*_i\|^2\right)\\
&\stackrel{(1)}\leq\sum_i\left(\opt_1(X^*_i)+2n_i\cdot\|\overline{c}_i-c^*_i\|^2+2n_i\cdot\|\hat{c}_i-\overline{c}_i\|^2\right)\\ %\textrm{\osnote{why 3 and not 2?}}\\
&\leq\sum_i\left(\opt_1(X^*_i)+2\opt_1(X^*_i)\cdot\frac{\rho}{1-\rho} + \right. \\
&\qquad \qquad \left. O(1)\cdot \min\left\{ n_i\cdot\Lambda^2,\; \frac{\Lambda^2 d}{\eps^2 (1-\rho)^2\cdot n_i}\cdot\ln^2\left(\frac{dk}{\beta\delta}\right) \right\} \right)\\
&\stackrel{(2)}\leq\sum_i\left(\opt_1(X^*_i)+2\opt_1(X^*_i)\cdot\frac{\rho}{1-\rho} +  O(\Lambda^2)\cdot  \frac{\sqrt{d}}{\eps (1-\rho)}\cdot\ln\left(\frac{dk}{\beta\delta}\right)  \right)\\
&=\left(1+\frac{2\rho}{1-\rho}\right)\cdot\opt_k(X) + O(1)\cdot  \frac{k \Lambda^2 \sqrt{d}}{\eps (1-\rho)}\cdot\ln\left(\frac{dk}{\beta\delta}\right)\\
&=\left(1+O(\phi^2)\right)\cdot\opt_k(X) + O(1)\cdot  \frac{k \Lambda^2 \sqrt{d}}{\eps}\cdot\ln\left(\frac{dk}{\beta\delta}\right).
\end{align*}
where Inequality $(1)$ follows from sum of squares bound (Claim~\ref{clm:sosBound}) and Inequality $(2)$ follows from the fact that $\min\left\{ n_i,\; \frac{d}{\eps^2 (1-\rho)^2\cdot n_i}\cdot\ln^2\left(\frac{dk}{\beta\delta}\right) \right\} \leq 2\frac{\sqrt{d}\ln\left(\frac{dk}{\beta\delta}\right)}{\eps (1-\rho)}$. Now, small clusters of size $n_i<\frac{16}{\eps(1-\rho)}\log(\frac{4k}{\beta\delta})$ can increase the cost of $\cost_X(\hat{C})$ by at most $\Lambda^2\cdot\frac{16}{\eps(1-\rho)}\log(\frac{4k}{\beta\delta})$ additively, and hence, overall we have that
$$
\cost_X(\hat{C})\leq \left(1+O(\phi^2)\right)\cdot\opt_k(X) + O(1)\cdot  \frac{k \Lambda^2 \sqrt{d}}{\eps}\cdot\ln\left(\frac{dk}{\beta\delta}\right).
$$
\end{proof}

Lemma~\ref{lem:goodC} shows that whenever the set of centers $B$ (computed in Step~1) is ``good enough'' then the resulting set of centers $\hat{C}$ has a low $k$-means cost (obtaining better guarantees than $B$). However, the set of centers $B$  is computed using a {\em private} approximation algorithm, which has both multiplicative and additive errors. In the next theorem we argue that, taking $B$'s additive error into account, either $B$ itself is already a good approximation for the $k$-means, or its additive error is small enough so that it has only a small effect on the error of $\hat{C}$.

\begin{theorem}\label{thm:main}
Let $X$ be a database containing $n$ points in the $d$-dimensional ball $\BBB(0,\Lambda)$, and assume that $X$ is $\phi$-separated for $k$-means. Let Algorithm \texttt{Private-Stable-$k$-Means} be executed on $X$ with a subroutine $\AAA$ that returns, with probability at least $(1-\beta_1)$, a set of centers $B$ satisfying $\cost_X(B)\leq v\cdot\opt_k(X)+t$. If $\phi^2\leq O\left(\frac{1}{v}\right)$, then with probability at least $(1-\beta-\beta_1)$, Algorithm \texttt{Private-Stable-$k$-Means} returns a set of centers $C'$ satisfying
$$
\cost_X(C')\leq(1+O(\phi^2))\cdot\opt_k(X)+O\left(vt+\frac{k \Lambda^2 \sqrt{d}}{\eps}\cdot\ln\left(\frac{dk}{\beta\delta}\right)\right).
$$
If furthermore $\opt_{k-1}(X)\geq \frac{t}{\phi^2}$, then
$$\cost_X(C')\leq(1+O(\phi^2))\cdot\opt_k(X)+O\left(\frac{k \Lambda^2 \sqrt{d}}{\eps}\cdot\ln\left(\frac{dk}{\beta\delta}\right)\right).$$
\end{theorem}

\begin{proof}[Proof sketch]
Recall that in Step~5, Algorithm \texttt{Private-Stable-$k$-Means} chooses between $B$ and $\hat{C}$ using the Gaussian mechanism. We analyze two cases and show that at least one of these options has small error (the additional error introduces by the Gaussian mechanism is also small). If $\opt_{k-1}(X)\leq \frac{t}{\phi^2}$ then
\begin{align*}
\cost_X(B)&\leq v\cdot\opt_k(X)+t\\
&\leq v\cdot\phi^2\cdot\opt_{k-1}(X)+t\\
&\leq v\cdot t+t = (v+1)\cdot t%\leq O(vt)\\
%&\leq(1+\phi^2)\cdot\opt_k(X)+O(t),
\end{align*}
and hence, $B$ is a good output. 
On the other hand, if $\opt_{k-1}(X)>\frac{t}{\phi^2}$ then
\begin{align*}
\cost_X(B)&\leq v\cdot\opt_k(X)+t\\
&\leq v\cdot\phi^2\cdot\opt_{k-1}(X)+\phi^2\cdot\opt_{k-1}(X)\\
&= \left(v+1\right)\phi^2\cdot\opt_{k-1}(X).
\end{align*}
Therefore, for $\phi^2\leq O(\frac{1}{v})$, we have that the conditions of Lemma~\ref{lem:goodC} are met, and so 
$$\cost_X(\hat{C})\leq \left(1+O(\phi^2)\right)\cdot\opt_k(X) + O(1)\cdot  \frac{k \Lambda^2 \sqrt{d}}{\eps}\cdot\ln\left(\frac{dk}{\beta\delta}\right),$$
and $\hat{C}$ is a good output. 
\end{proof}

Combining Theorem~\ref{thm:main} with the private algorithm of~\cite{KaplanSt18} achieving $O(1)$-approximation for the $k$-means, we get the following corollary.\footnote{For simplicity, throughout the paper we use the $\tilde{O}$ notation to hide logarithmic factors in $k,n,d,\beta,\delta$.}

\begin{corollary}\label{cor:main}
There exists an $(\eps,\delta)$-differentially private algorithm such that the following holds. 
Let $X$ be a database containing $n$ points in the $d$-dimensional ball $\BBB(0,\Lambda)$, and assume that $X$ is $\phi$-separated for $k$-means for $\phi=O(1)$ (sufficiently small). When applied to $X$, the algorithm returns, with probability at least $(1-\beta)$, a set of $k$ centers $C'$ satisfying
$$
\cost_X(C')\leq(1+O(\phi^2))\cdot\opt_k(X)+\tilde{O}\left(\frac{k^{1.01}\cdot d^{0.51}\cdot\Lambda^2}{\eps^{1.01}}+\frac{k^{1.5}\cdot\Lambda^2}{\eps}\right).
$$
If furthermore $\opt_{k-1}(X)\geq \tilde{O}\left(\frac{k^{1.01}\cdot d^{0.51}\cdot\Lambda^2}{\eps^{1.01}\phi^2}+\frac{k^{1.5}\cdot\Lambda^2}{\eps\phi^2}\right)$, then
$$\cost_X(C')\leq(1+O(\phi^2))\cdot\opt_k(X)+O\left(\frac{k \Lambda^2 \sqrt{d}}{\eps}\cdot\ln\left(\frac{dk}{\beta\delta}\right)\right).$$
\end{corollary}

As we mentioned,~\cite{NRS07,Wang2015,HuangL18} also presented private algorithms for $k$-means on well-separated instances, based on the sample and aggregate technique. Their constructions, however, only guarantee approximation in terms of the {\em Wasserstein distance} to the optimal centers. In Section~\ref{sec:alternative} we show that small modifications to the algorithm of Huang and Liu~\cite{HuangL18}, together with a refined analysis, yields an alternative algorithm for privately approximating the $k$-means in terms of {\em $k$-means cost}. While the error bound we get in Section~\ref{sec:alternative} is strictly weaker than our construction from Corollary~\ref{cor:main}, we believe that the alternative construction of Section~\ref{sec:alternative} could still be of interest for future works, as it involves different ideas than those used for obtaining Corollary~\ref{cor:main}.

Our Algorithm \texttt{Private-Stable-$k$-Means} also results in a new construction for privately approximating the $k$-means in terms of the Wasserstein distance to the optimal centers. This follows from the fact that, for well-separated instances, centers with near optimal $k$-means cost must be close to the optimal centers in terms of the Wasserstein distance. Specifically, we can use the following theorem.

\begin{theorem}[{\cite[Lemma 3.3]{HuangL18}}]\label{thm:HuangL} 
Let $X$ be a database containing $n$ points from the $d$-dimensional ball $\BBB(0,\Lambda)$, and assume that $X$ is $\phi$-separated for $k$-means for $\phi$ sufficiently small. Let $C^*=\{c^*_1,\dots,c^*_k\}$ denote a set of optimal centers for $X$, and let $\hat{C}=\{\hat{c}_1,\dots,\hat{c}_k\}$ be a set of centers such that $\cost_X(\hat{C})\leq\opt_{k}(X)+s$ and $\cost_X(\hat{C})\leq\frac{1}{800}\opt_{k-1}(X)$. 
Furthermore, assume that the centers in $\hat{C}$ are the result of a Lloyd step, i.e., these centers are obtained by averaging the corresponding clusters in $X$.
 Then, the Wasserstein distance between $C^*$ and $\hat{C}$ is at most $O\left(\Lambda\cdot\left(\phi^2+\sqrt{\frac{s}{\opt_{k-1}(X)}}\right)\right)$.
\end{theorem}

Instantiating Theorem~\ref{thm:HuangL} with our k-means algorithm for the centralized model (Corollary~\ref{cor:main}), we get the following result.\footnote{Theorem~\ref{thm:HuangL} requires the centers $\hat{C}$ to be obtained by averaging the corresponding clusters of $X$, while our algorithm only averages some of the points within each cluster. Formally, the requirement of Theorem~\ref{thm:HuangL} can be met by adding an additional (noisy) Lloyd step at the end of our algorithm (the noise introduced for privacy in this additional Lloyd step is of a lower order).}

\begin{theorem}\label{thm:wd}
There exists an $(\eps,\delta)$-differentially private algorithm such that the following holds. 
Let $X$ be a database containing $n$ points in the $d$-dimensional ball $\BBB(0,\Lambda)$. %\osnote{Confused, do you mean $\BBB(0,\Lambda)$?}
Assume that $X$ is $\phi$-separated for $k$-means for $\phi=O(1)$ (sufficiently small), and assume that
$$\opt_{k-1}(X)\geq\tilde{O}\left(
\left(\frac{\Lambda^2}{\phi^4} \right)\cdot
\left( \frac{k^{1.01}\cdot d^{0.51}}{\eps^{1.01}}+
\frac{k^{1.5}}{\eps}
\right)
\right).$$
When applied to $X$, the algorithm returns, with probability at least $(1-\beta)$, a set of $k$ centers $C'$ satisfying $\wdist(C^*,C')\leq O(\phi^2\cdot\Lambda)$, where $C^*$ are the optimal centers.
\end{theorem}

%As Huang and Liu~\cite{HuangL18} showed, Wasserstein distance of $O(\phi^2)$ is the best possible under differential privacy, and hence, the approximation factor in Theorem~\ref{thm:wd} is tight. 
%Huang and Liu also presented an algorithm that outputs a center set that is within a Wasserstein distance of $O(\phi^2\cdot\Lambda)$ from the optimal center set. However, in their construction, this bound is only guaranteed to hold whenever $$\opt_{k-1}(X)\geq \tilde{O}\left(n^{\frac{11}{20}}k^{\frac{7}{4}}d^{\frac{3}{4}}\Lambda^2\eps^{-\frac{1}{2}}\phi^{-4}\right),$$ whereas our bound is guaranteed to hold even for much smaller values of $\opt_{k-1}(X)$. In particular, in our construction $\opt_{k-1}(X)$ dependency on $n$ is only poly-logarithmic.
The error bound in this theorem matches the state-of-the-art result of \cite{HuangL18}, and offers some improvements in terms of the requirement on $\opt_{k-1}(X)$. Specifically, the bound of \cite{HuangL18} is guaranteed to hold whenever $\opt_{k-1}(X)\gtrsim n^{\frac{11}{20}}k^{\frac{7}{4}}d^{\frac{3}{4}}\eps^{-\frac{1}{2}}\phi^{-4}$, whereas our bound holds also for smaller values of $\opt_{k-1}(X)$. In particular, in our construction $\opt_{k-1}(X)$ dependency on $n$ is only poly-logarithmic.

%%%%%%%%%%%%%%%%%%%%%%%%%%%%%%%%%%%%%%%%%%
% $k$-median with stability assumption
%%%%%%%%%%%%%%%%%%%%%%%%%%%%%%%%%%%%%%%%%%
\newpage
\section{Private $k$-median Clustering with Stability Assumptions}\label{sec:median}
Our construction for the $k$-median is conceptually similar to our construction for the $k$-means. Specifically, we first apply a private $k$-median approximation algorithm on the data to obtain $k$ centers $B=\{b_1,\dots,b_k\}$, use these centers to partition the data into $k$ clusters $X_{b_1},\dots,X_{b_k}$, and then privately compute an appropriate center for each cluster $X_{b_i}$. The main difference is that in Section~\ref{sec:main} we could privately compute a center for each $X_{b_i}$ as a noisy average (using the Gaussian mechanism). For $k$-median, however, using the average to compute the center of each cluster is not a good option, as it can be far from the optimal median of the cluster. We overcome this issue by replacing the Gaussian mechanism with a tool of Bassily et al.~\cite{BassilyST14} for privately solving convex optimization problems. 
%While for $k>1$ the $k$-median cost objective is not convex, for $k=1$ it is convex, which is exactly what we need in our construction, as we have already partitioned the data into $k$ different clusters (using the centers in $B$).
Once we $k$-partition the data (using the centers in $B$) we then use private stochastic gradient descent to approximate the $1$-median center of each subset in the partition. 

\begin{theorem}[Bassily et al.~\cite{BassilyST14}]\label{thm:BST}
Let $\XXX$ be an arbitrary domain, let $\CCC\subseteq\R^d$ be a  closed and convex set with diameter $\|\CCC\|$, and let $\ell:\R^d\times\XXX\rightarrow\R$ be such that $\ell(\cdot,x)$ is convex and $L$-Lipschitz for all $x\in\XXX$. There exists an $(\eps,\delta)$-differentially private algorithm that takes a database $X=(x_1,\dots,x_n)\in\XXX^n$ and returns a value $\hat{w}\in\R^d$ s.t.\ with probability at least $1-\beta$ we have
$$
\sum_{i=1}^n \ell(\hat{w},x_i) - \min_{w\in\CCC\subseteq\R^d} \sum_{i=1}^n \ell(w,x_i) \leq \frac{\sqrt{d}\cdot L\cdot\|\CCC\|}{\eps}\cdot\polylog\left(n,\frac{1}{\beta},\frac{1}{\delta}\right).
$$ 
\end{theorem}

In particular, for $\CCC=\BBB(0,\Lambda)$ and $\ell(w,x)=\|w-x\|$, which is convex and 1-Lipschitz, we can use the above theorem to identify an approximate median of the database $X$, with additive error at most $\approx\frac{\Lambda\sqrt{d}}{\eps}$. For simplicity, we assume (without loss of generality) that the algorithm from Theorem~\ref{thm:BST} is differentially private w.r.t.\ adding/removing an element from the database. Our construction for $k$-median appears in Algorithm \texttt{Private-Stable-$k$-median}.

The privacy properties of Algorithm \texttt{Private-Stable-$k$-median} are straight forward (follow from composition, see~\cite{dwork2010boosting}). Before proceeding with the utility analysis, we restate Theorem~\ref{thm:ostKmeans} for the case of $k$-median (instead of $k$-means), a result required for showing that the centers in $B$ (computed in Step~1) are close to the optimal centers. As before, the optimal $k$-median centers are denoted by $c_1,..,c_k$ and for each $i$ we denote $D_i = \min_{j\neq i}\|c_i-c_j\|$. The proof of this theorem is deferred to Appendix~\ref{apx_sec:ostrovskythmproof}.

\begin{theorem}[{{\cite[Theorem 5.1]{DBLP:journals/jacm/OstrovskyRSS12} for $k$-median}}]\label{thm:ostMedian}
Let $\alpha$ and $\phi$ be such that $\frac{\alpha+\phi}{1-\phi}<\frac{1}{4}$. Suppose that $X\subseteq\R^d$ is $\phi$-separated for $k$-median, let $C=(c_1,\dots,c_k)$ be a set of optimal centers for $X$, and let $\hat{C}=(\hat{c}_1\dots,\hat{c}_k)$ be centers such that $\medcost_X(\hat{C})\leq\alpha\cdot\medopt_{k-1}(X)$. Then for each $\hat{c}_i$ there is a distinct optimal center, call it $c_i$, such that $\|\hat{c}_i-c_i\|\leq2\cdot\frac{\alpha+\phi}{1-\phi}\cdot D_i$, where $D_i = \min_{j\neq i}\|c_i-c_j\|$.
\end{theorem}

\begin{algorithm*}[t]

\caption{\texttt{Private-Stable-$k$-median}}\label{alg:privateKmedian}

{\bf Input:} Database $X$ containing $n$ points in the $d$-dimensional ball $\BBB(0,\Lambda)$, failure probability $\beta$, privacy parameters $\eps,\delta$.

{\bf Tool used:} An $(\eps,\delta)$-differentially private algorithm $\AAA$ for approximating the $k$-median.

\begin{enumerate}[leftmargin=15pt,rightmargin=10pt,itemsep=1pt,topsep=0pt]

\item Run $\AAA$ on $X$ to obtain $k$ centers: $B=\{b_1,\dots,b_k\}$.

\item Let $X_{b_1},\dots,X_{b_k}\subseteq X$ be the partition of the inputs points according to the centers $B$. That is, $X_{b_i}=\{x\in X : i=\argmin_j\|x-b_j\|\}$.

\item For $i\in[k]$ use the algorithm from Theorem~\ref{thm:BST} with privacy parameters $\eps,\delta$ and confidence parameter $\frac{\beta}{k}$ to identify an approximate 1-median $\hat{c}_i$ of $X_{b_i}$, that is $\medcost_{X_{b_i}}(\{\hat{c}_i\})\approx\medopt_k(X_{b_i})$. Denote $\hat{C}=\{\hat{c}_1,\dots,\hat{c}_k\}$.

\item Use the Gaussian mechanism with privacy parameters $(\eps,\delta)$ to estimate $\medcost_X(\hat{C})$ and $\medcost_X(B)$. Output the set of centers (either $\hat{C}$ or $B$) with the lower (estimated) cost.

\end{enumerate}
\end{algorithm*}

We are now ready to present the utility analysis of \texttt{Private-Stable-$k$-median}. The main ingredient in this analysis is captured by the following lemma, in which we analyze the cost of $\hat{C}$ (the set of centers computed in Step~3 of the execution).

\begin{lemma}\label{lem:mainMed}
Let $X$ be $\phi$-separated for $k$-median for $\phi$ sufficiently small, and consider the execution of \texttt{Private-Stable-$k$-median} on $X$. Let $B$ and $\hat{C}$ denote the centers from Steps~1 and~3. If $\medcost_X(B)\leq w\cdot\medopt_{k}(X)$, then 
$$\medcost_X(\hat{C})\leq\left(1+\frac{8(w+1)\phi}{1-(8w+9)\phi}\right)\cdot\medopt_{k}(X) + \tilde{O}\left(\frac{k\Lambda\sqrt{d}}{\eps} \right).$$
\end{lemma}
%$$\medcost_X(\hat{C})\leq\left(1+\frac{8(w+1)\phi}{1-(8w+9)\phi}\right)\cdot\medopt_{k}(X) + O\left(\frac{k\Lambda\sqrt{d}}{\eps}\cdot\ln\left(\frac{kd}{\beta\delta}\right)\right).$$

\begin{proof}
%Let $X$ be a database with optimal centers $C^*=\{c^*_1,\dots,c^*_k\}$, and consider the execution of \texttt{Private-Stable-$k$-Means} on $X$.  
%
For a given set of centers $A$ and a point $x$ we write $A(x)$ to denote the nearest neighbor of $x$ in $A$. 
Let $C^*=\{c^*_1,\dots,c^*_k\}$ be a set of optimal centers for $X$, and consider the following feasible (but not necessarily optimal) assignment of the points in $X$ to the centers $C^*$: instead of assigning a point $x\in X$ to its nearest neighbor in $C^*$, we assign it to $C^*(B(x))$. That is, to assign the point $x$ to a center we first find its nearest neighbor in $B$, call it $b$, and the assign $x$ to the nearest neighbor of $b$ in $C^*$. Let us denote the $k$-median cost of this assignment as $\medcost_{X\rightarrow B}(C^*)=\sum_{x\in X}\|x-C^*(B(x))\|$. 

We want to compare $\medcost_{X\rightarrow B}(C^*)$ with $\medcost_X(C^*)=\medopt_k(X)$. Observe that if for a point $x\in X$ we have that $C^*(x)=C^*(B(x))$, then the cost of this point remains the same in both assignments. Now let $x\in X$ be such that $C^*(x)\neq C^*(B(x))$. We have that
\begin{align}
\|x-C^*(x)\|+\|C^*(x)-B(C^*(x))\|&\geq \|x-B(C^*(x))\|\nonumber\\
&\geq\|x-B(x)\|\nonumber\\
&\geq\|x-C^*(B(x))\| - \|B(x)-C^*(B(x))\|\label{eq:lem:mainMed_eq1}
\end{align}
Recall that $\medcost_X(B)\leq w\cdot\medopt_{k}(X)\leq w\cdot\phi\cdot\medopt_{k-1}(X)$.
Denote $\gamma=2\frac{(w+1)\cdot\phi}{1-\phi}$, and $D=\|C^*(x)-C^*(B(x))\|$. By Theorem~\ref{thm:ostMedian} we have that 
$$
\|C^*(x)-B(C^*(x))\|\leq\gamma\cdot D \qquad \text{and} \qquad \|B(x)-C^*(B(x))\|\leq\gamma\cdot D.
$$
Together with Inequality~(\ref{eq:lem:mainMed_eq1}) this means that
\begin{align}
\|x-C^*(B(x))\|&\leq \|x-C^*(x)\| + 2\gamma\cdot D.\label{eq:lem:mainMed_eq2}
\end{align}
Next observe that
\begin{align*}
D&=\|C^*(x)-C^*(B(x))\|\\
&\leq \|C^*(x)-B(C^*(x))\|+\|B(C^*(x))-x\|+\|x-B(x)\|+\|B(x)-C^*(B(x))\|\\
&\leq 2\gamma D + \|B(C^*(x))-x\|+\|x-B(x)\|\\
&\leq 2\gamma D + 2\cdot\|B(C^*(x))-x\|\\
&\leq 2\gamma D + 2\left(\|B(C^*(x))-C^*(x)\|+\|C^*(x)-x\|\right)\\
&\leq 2\gamma D + 2\left(\gamma D+\|C^*(x)-x\|\right) = 4\gamma D + 2\|C^*(x)-x\|
\end{align*}
which means that $D\leq\frac{2}{1-4\gamma}\cdot\|C^*(x)-x\|$. Together with inequality~(\ref{eq:lem:mainMed_eq2}) we get that
\begin{align*}
\|x-C^*(B(x))\|&\leq \left(1+\frac{4\gamma}{1-4\gamma}\right)\cdot \|x-C^*(x)\|.
\end{align*}
Hence,
\begin{align*}
\medcost_{X\rightarrow B}(C^*)&=\sum_{x\in X}\|x-C^*(B(x))\|\\
&\leq\sum_{x\in X}\left(1+\frac{4\gamma}{1-4\gamma}\right)\cdot \|x-C^*(x)\|\\
&\leq\left(1+\frac{4\gamma}{1-4\gamma}\right)\cdot\medcost_X(C^*)=\left(1+\frac{4\gamma}{1-4\gamma}\right)\cdot\medopt_k(X).
\end{align*}
The above inequality allows us to relate $\medopt_k(X)$ to $\medcost_{X\rightarrow B}(C^*)$. To relate $\medcost_{X\rightarrow B}(C^*)$ to the cost of the output centers $\hat{C}$, recall that each center $\hat{c}_{i}$ is obtained by computing an approximate median of $X_{b_i}$, with small additive error. By Theorem~\ref{thm:BST} and by a union bound over $i\in[k]$, with probability at least $(1-\beta)$, for all $i\in[k]$ we have that 
$\medcost_{X_{b_i}}(\{\hat{c}_i\})\leq\medcost_{X_{b_i}}(\Med(X_{b_i}))+\tilde{O}\left(\frac{\sqrt{d}\Lambda}{\eps}\right)$, where $\Med(X_{b_i})$ minimizes the $1$-median cost of $X_{b_i}$. Therefore, 
\begin{align*}
\medcost_X(\hat{C})&\leq\sum_{i\in[k]}\sum_{x\in X_{b_i}}\|x-\hat{c}_i\|\\
&\leq\sum_{i\in[k]}\left(\tilde{O}\left(\frac{\sqrt{d}\Lambda}{\eps}\right)+\sum_{x\in X_{b_i}}\|x-\Med(X_{b_i})\|\right)\\
&=\tilde{O}\left(\frac{k\sqrt{d}\Lambda}{\eps}\right)+\sum_{i\in[k]}\sum_{x\in X_{b_i}}\|x-\Med(X_{b_i})\|\\
&\leq\tilde{O}\left(\frac{k\sqrt{d}\Lambda}{\eps}\right)+\sum_{i\in[k]}\sum_{x\in X_{b_i}}\|x-C^*(B(x))\|\\
&=\tilde{O}\left(\frac{k\sqrt{d}\Lambda}{\eps}\right)+\sum_{x\in X}\|x-C^*(B(x))\|\\
&\leq\tilde{O}\left(\frac{k\sqrt{d}\Lambda}{\eps}\right)+\left(1+\frac{4\gamma}{1-4\gamma}\right)\cdot\medopt_k(X).
\end{align*}
\end{proof}

%\osnote{You must add more to this. The work gives a $O(1)$-approx of the $k$-means. It is applicable for the $k$-median, but the uninformed reader will not know this. You have to expand here substantially.}
Similarly to the analysis of Theorem~\ref{thm:main}, combining Lemma~\ref{lem:mainMed} with the private algorithm of~\cite{KaplanSt18}, that achieves $O(1)$-approximation for the $k$-median, yields the following result.\footnote{Kaplan and Stemmer~\cite{KaplanSt18} stated their result only for $k$-means, but their construction carries over to $k$-median with almost no modifications.}

\begin{theorem}
There exists an $(\eps,\delta)$-differentially private algorithm such that the following holds. 
Let $X$ be a database containing $n$ points in the $d$-dimensional ball $\BBB(0,\Lambda)$, and assume that $X$ is $\phi$-separated for $k$-median for $\phi=O(1)$ (sufficiently small). When applied to $X$, the algorithm returns, with probability at least $(1-\beta)$, a set of $k$ centers $C'$ satisfying
$$
\medcost_X(C')\leq(1+O(\phi))\cdot\medopt_k(X)+\tilde{O}\left(\frac{k^{1.01}\cdot d^{0.51}\cdot\Lambda}{\eps^{1.01}}+\frac{k^{1.5}\cdot\Lambda}{\eps}\right).
$$
\end{theorem}

%%%%%%%%%%%%%%%%%%%%%%%%%%%%%%%%%%%%%%%%%%%%%%%%%%%%%%%%%%
% $k$-means with stability assumption for the local model
%%%%%%%%%%%%%%%%%%%%%%%%%%%%%%%%%%%%%%%%%%%%%%%%%%%%%%%%%%
\newpage
\section{Clustering with Stability Assumptions in the Local Model}
\label{sec:kMeansStabilityLocal}

\subsection{Additional preliminaries}
We now present additional preliminaries from local differential privacy that enable our construction. The local model of differential privacy was formally defined first in~\cite{KLNRS08}. We give here the formulation presented by Vadhan~\cite{Vadhan2016}.

Consider $n$ parties $P_1,\dots,P_n$, where each party is holding a data item $x_i$. We denote $X=(x_1,\dots,x_n)$ and refer to $X$ as a {\em distributed database}. A {\em protocol} proceeds in a sequence of rounds until all (honest) parties terminate. Informally, in each round, each party selects a message to be broadcast based on its input, internal coin tosses, and all messages received in previous rounds. The {\em output} of the protocol is specified by a deterministic function of the transcript of messages exchanged. 

For some $j\in[n]$, we consider an {\em adversary} controlling all parties other than $P_j$. Given a particular adversary strategy $A$, we write $\view_{A}((A\leftrightarrow(P_1,\dots,P_n))(X))$ for the random variable that includes everything that $A$ sees when participating in the protocol $(P_1,\dots,P_n)$ on input $X=(x_1,\dots,x_n)$.

\begin{definition}[Local differential privacy~\cite{KLNRS08,BeimelNO08,DR14,Vadhan2016}]
A protocol $P=(P_1,\dots,P_n)$ satisfies $(\eps,\delta)$-local differential privacy (LDP) if, for every $j\in[n]$, for every adversary $A$ controlling all parties other than $P_j$, for every two datasets $X,X'$ that differ on $P_j$'s input (and are equal otherwise), the following holds for every set $T$:
$$
\Pr[\view_{A}((A\leftrightarrow(P_1,\dots,P_n))(X))\in T]
\leq e^{\eps}\cdot\Pr[\view_{A}((A\leftrightarrow(P_1,\dots,P_n))(X'))\in T]+\delta.
$$
\end{definition}

As is standard in the literature on local differential privacy, we consider protocols in which there is a unique player, called {\em the server}, which has no input of its own. All other players are called {\em users}. Typically, users do not communicate with other users, only with the server.

\paragraph{Counting queries and histograms with local differential privacy.} 
The most basic task that we can apply in the local differential privacy model is {\em counting}. Let $X\in \{0,1\}^n$ be a database which is distributed among $n$ users (each holding one bit), and consider the task of estimating the number of users holding a $1$. This can be solved privately with error proportional to $\frac{1}{\eps}\sqrt{n}$ (see, e.g.,~\cite{KLNRS08}). A more general setting is when instead of a binary domain, every user holds an input item from some (potentially) large domain $U$. This can be solved using tools from the recent line of work on heavy hitters in the local model.~\cite{HsuKR12,BassilyS15,BNST17,BunNS18}

\paragraph{Notation.} For a database $X=(x_1,\dots,x_n)\in U^n$ and a domain element $u\in U$, we use $f_X(u)$ to denote the multiplicity of $u$ in $X$, i.e., $f_X(u)=|\{x_i\in X : x_i=u\}|.$

\begin{theorem}[\cite{HsuKR12,BassilyS15,BNST17,BunNS18}]\label{thm:HH}
Fix $\beta,\eps\leq1$. There exists an $(\eps,0)$-LDP protocol that operates on a (distributed) database $X\in U^n$
for some finite set $U$, and returns a mapping $\hat{f}:U\rightarrow\R$ such that the following holds. For every choice of $u\in U$, with probability at least $1-\beta$, we have that
$$\left|\hat{f}(u) - f_X(u)\right|\leq \frac{3}{\eps}\cdot\sqrt{n\cdot \log\left(\frac{4}{\beta}\right)}.$$
%$$\left|\hat{f}(u) - f_X(u)\right|\leq O\left(\frac{1}{\eps}\cdot\sqrt{n\cdot \log\left(\frac{n}{\beta}\right)}\right).$$
\end{theorem}

\paragraph{Average of vectors in $\R^d$.} 
Consider a (distributed) database $X=(x_1,\dots,x_n)$ where every user $i$ is holding $x_i\in\R^d$. One of the most basic tasks we can apply under local differential privacy is to compute a noisy estimation for the sum (or the average) of vectors in $X$. Specifically, every user sends the server a noisy estimation of its vector (e.g., by adding independent Gaussian noise to each coordinate), and the server simply sums all of the noisy reports to obtain an estimation for the sum of $X$.

\begin{theorem}[folklore]\label{thm:LDPsum}
Consider a (distributed) database $X=(x_1,\dots,x_n)$ where every user $i$ is holding a point $x_i$ in the $d$ dimensional ball $\BBB(0,\Lambda)$. There exists an $(\eps,\delta)$-LDP protocol for computing an estimation $\vec{a}$ for the sum of the vectors in $X$, such that with probability at least $(1-\beta)$ we have
$$
\left\|\vec{a}-\sum_{i\in[n]}x_i\right\|\leq\frac{2\Lambda\sqrt{nd}\ln(\frac{2}{\beta\delta})}{\eps}.
$$
\end{theorem}

For our constructions we need a tool for computing averages of {\em subsets} of $X$. Specifically, assume that there are $n$ users, where user $i$ is holding a point $x_i\in\BBB(0,\Lambda)$. Moreover, assume that we have a fixed (publicly known) partition of $\BBB(0,\Lambda)$ into a finite number of regions: $R_1,\dots,R_T\subseteq\BBB(0,\Lambda)$. For every region $R_{\ell}$, we would like to obtain an estimation for the average of the input points in that region. For this purpose we will use the following simple protocol, called \texttt{LDP-AVG} (for an analysis see, e.g.,~\cite{Stemmer20}).

{\floatname{algorithm}{Protocol}
\begin{algorithm*}[h!]
\caption{\texttt{LDP-AVG}}

\noindent {\bf Public parameters:} Partition of the $d$-dimensional ball $\BBB(0,\Lambda)$ into $T$ regions $R_1,\dots,R_T$.

\smallskip

\noindent {\bf Setting: }Each user $i\in[n]$ holds a point $x_i\in \BBB(0,\Lambda)$. Define $X=(x_1,\dots,x_n)$.

\begin{enumerate}[leftmargin=15pt,rightmargin=10pt,itemsep=1pt]

\item[{\bf 1.}]{\bf Every user $\boldsymbol{i}$:} Let $y_i\in(\R^d)^T$ be a vector whose every coordinate is sampled i.i.d.\ from $\NNN(0,\sigma^2)$, for $\sigma=\frac{8\Lambda}{\eps}\sqrt{\ln(1.25/\delta)}$. Let $t$ be s.t.\ $x_i\in R_t$. Add $x_i$ to $y_{i,t}$. Send $y_i$ to the server.

\item[{\bf 2.}]{\bf The server and the users:} Run the protocol from Theorem~\ref{thm:HH} with privacy parameter $\frac{\eps}{2}$. For every $t\in[T]$ the server obtains an estimation $\hat{r}_t\approx|\{i:x_i\in R_t\}|\triangleq r_t$.

\item[{\bf 3.}]{\bf The server:} Output a vector $\hat{a}\in(\R^d)^T$, where $\hat{a}_t=\frac{1}{\hat{r}_t}\cdot\sum_{i\in[n]}y_{i,t}$.
\end{enumerate}
\end{algorithm*}}

\begin{Claim}
\texttt{LDP-AVG} satisfies $(\eps,\delta)$-LDP. Moreover, with probability at least $(1-\beta)$, for every $t\in[T]$ s.t.\ $r_t\geq\frac{12}{\eps}\cdot\sqrt{n\cdot \log\left(\frac{4T}{\beta}\right)}$ we have that 
$$
\left\|\frac{1}{\hat{r}_t}\cdot\sum_{i\in[n]}y_{i,t}-\frac{1}{r_t}\cdot\sum_{\substack{i\in[n]:\\x_i\in R_t}}x_i\right\|\leq\frac{48\sqrt{dn}\Lambda\cdot\ln(\frac{8dT}{\beta\delta})}{\eps\cdot r_t}.
$$
\end{Claim}

\subsection{A Locally-Private Clustering Algorithm for Well-Separated Instances}
\label{subsec:LDP-k-means-niceinput}

{\floatname{algorithm}{Protocol}
\begin{algorithm*}[t]

\caption{\texttt{LDP-Stable-$k$-Means}}\label{alg:LDPkMeans}

{\bf Input:} Failure probability $\beta$, privacy parameters $\eps,\delta$.

\smallskip
\noindent {\bf Setting: }Each player $i\in[n]$ holds a point $x_i$ in the $d$-dimensional ball $\BBB(0,\Lambda)$. Define $X=(x_1,\dots,x_n)$.

\smallskip
\noindent {\bf Tool used: }An $(\eps,\delta)$-LDP protocol $\AAA$ for approximating the $k$-means.

\begin{enumerate}[leftmargin=15pt,rightmargin=10pt,itemsep=1pt]

\item Run $\AAA$ on $X$ to obtain $k$ centers: $B=\{b_1,\dots,b_k\}$.

\item For $i\in[k]$ let $\hat{D}_i=\min_{j\neq i}\|b_i-b_j\|$.

\item For $i\in[k]$ let $R_i=\{x\in\BBB(0,\Lambda) : \|x-b_i\|\leq\hat{D}_i/3\}$, and denote $\hat{X}_i=X\cap R_i$.

\item Let $\overline{C}=\{\overline{c}_1,\dots,\overline{c}_k\}$ denote the average of the points in $\hat{X}_1,\dots,\hat{X}_k$, respectively. Use \texttt{LDP-AVG} with privacy parameters $(\eps,\delta)$ to obtain for every $i\in[k]$ a noisy estimation $\hat{c}_i$ of the average of $X\cap R_i$, i.e., an estimation of $\overline{c}_i$.

\item Estimate $\cost_X(\hat{C})$ and $\cost_X(B)$, e.g., using Theorem~\ref{thm:LDPsum}. Output the set of centers (either $\hat{C}$ or $B$) with the lower estimated cost.

\end{enumerate}
\end{algorithm*}}

Our protocol for the local model is obtained from our construction for the centralized model by using \texttt{LDP-AVG} to compute averages instead of the Gaussian mechanism. The full construction appears in protocol \texttt{LDP-Stable-$k$-Means}.
A similar analysis to that of Section~\ref{sec:main} shows the following theorem.

\begin{theorem}\label{thm:mainLDP}
Let \texttt{LDP-Stable-$k$-Means} be executed with a subroutine $\AAA$ that returns, with probability at least $(1-\beta_1)$, a set of centers $B$ satisfying $\cost_X(B)\leq v\cdot\opt_k(X)+t$. If $\phi^2\leq O\left(\frac{1}{v}\right)$, then with probability at least $(1-\beta-\beta_1)$, protocol \texttt{LDP-Stable-$k$-Means} returns a set of centers $C'$ satisfying
$$
\cost_X(C')\leq(1+O(\phi^2))\cdot\opt_k(X)+O\left(vt+\frac{k \Lambda^2 \sqrt{dn}}{\eps}\cdot\ln\left(\frac{dk}{\beta\delta}\right)\right).
$$
\end{theorem}

Combining Theorem~\ref{thm:mainLDP} with the locally private protocol of~\cite{KaplanSt18,Stemmer20} achieving $O(1)$-approximation for the $k$-means, we get the following corollary.%\footnote{\osnote{REALLY redundant, wouldn't you say again??}For simplicity, we use the $\tilde{O}$ notation to hide logarithmic factors in $k,n,d,\beta,\delta$.}

\begin{corollary}\label{cor:mainLDP}
There exists an $(\eps,\delta)$-LDP protocol such that the following holds. 
Let $X$ be a (distributed) database containing $n$ points in the $d$-dimensional ball $\BBB(0,\Lambda)$, and assume that $X$ is $\phi$-separated for $k$-means for $\phi=O(1)$ (sufficiently small). When applied to $X$, the protocol returns, with probability at least $(1-\beta)$, a set of $k$ centers $C'$ satisfying
$$
\cost_X(C')\leq(1+O(\phi^2))\cdot\opt_k(X)+\tilde{O}\left(\frac{k\Lambda^2}{\eps}\cdot n^{0.51} \cdot \sqrt{d}\right).
$$
\end{corollary}

A similar result holds also for $k$-median, by instantiating the locally-private tool of Feldman et al.~\cite{FeldmanMTT18} for solving convex optimization problems. As in the centralized model, using Theorem~\ref{thm:HuangL}, our LDP protocol for the $k$-means gives also a good approximation of the optimal centers in terms of the Wasserstein distance. 
%\osnote{The following sentence belongs in the intro, not here.}This is the first LDP protocol for $k$-means clustering w.r.t.\ the Wasserstein distance. 
Specifically,
%Instantiating Theorem~\ref{thm:HuangL} with our $k$-means protocol for the local model (see Corollary~\ref{cor:mainLDP}), we get the following result.

\begin{theorem}\label{thm:wdLDP}
There exists an $(\eps,\delta)$-LDP protocol such that the following holds. 
Let $X$ be a (distributed) database containing $n$ points in the $d$-dimensional ball $\BBB(0,\Lambda)$. %\osnote{Again, not $\BBB(0,\Lambda)$? But now the equation below has no $\Lambda$...}. 
Assume that $X$ is $\phi$-separated for $k$-means for sufficiently small $\phi$,
and assume that 
%$$\opt_{k-1}(X)\geq \tilde{O}\left(\frac{\Lambda^2}{\eps}\cdot n^{0.67} \cdot d^{1/3}+\frac{\Lambda^2}{\phi^2\cdot\eps^{1/2}}\cdot n^{0.34}\cdot d^{1/6}\right).$$
\begin{align*}
&\opt_{k-1}(X)\geq \tilde{O}\left(\frac{k \sqrt{d}\cdot n^{0.51}\cdot\Lambda^2}{\eps\cdot\phi^4}  \right).
\end{align*}
Then on $X$, the protocol returns with probability $\geq (1-\beta)$ a set of $k$ centers $C'$ satisfying $\wdist(C^*,C')\leq O(\phi^2\cdot\Lambda)$, with $C^*$ denoting the optimal centers.
\end{theorem}

%%%%%%%%%%%%%%%%%%%%%%%%%%%%%%%%%%%%%%%%%%%%%%%%%%%
% Sample and Aggregate structure k-means algorithm
% (An alternative construction for k-means) 
%%%%%%%%%%%%%%%%%%%%%%%%%%%%%%%%%%%%%%%%%%%%%%%%%%%
\newpage
\section{An Alternative Construction Based on Sample and Aggregate}\label{sec:alternative}

In this section we present a variation of the sample-and-aggregate based approach of~\cite{HuangL18} for the problem of stable $k$-means clustering. In a nutshell, just like in~\cite{HuangL18} we partition the data into $T$ subsets, apply a non-private clustering algorithm to each subset to get $k$ useful centers per subset, and then apply the $1$-cluster algorithm~\cite{NS18_1Cluster} repeatedly to the $Tk$ resulting datapoints to retrieve $k$ centers. However, as opposed to~\cite{HuangL18}, we conclude with a Lloyd step over the resulting $k$ centers rather then applying yet again the $1$-cluster algorithm. The bulk of the analysis in this section is similar to the analysis of Huang and Liu~\cite{HuangL18}.%\cite{NS18_1Cluster}

Although the results in this section are incremental, the contribution of the sample-and-aggregate framework to our work lies in its conceptual significance. The algorithm presented in Section~\ref{sec:main} ties together $v$, the approximation guarantee of the DP algorithm \emph{in the worst-case}, with $\phi^2$, the quality of the guarantee of the input's ``niceness'' which differentiates it from a worst-case instance. (In particular, a necessary condition for its applicability is that $\phi^2\leq O(1/v)$.) In contrast, the sample-and-aggregate based approach is devoid of such a requirement and is therefore potentially applicable under a wider range of $\phi^2$-values. More importantly, it severs the tie between worst-case approximation guarantee and the input-niceness guarantee. In other words, if we wish to apply the algorithm from Section~\ref{sec:main} and obtain a good approximation for an instance with a moderate level of separability, it is required we improve upon the $k$-means approximation guarantee in the worst-case; whereas the sample-and-aggregate framework gives a meaningful guarantee of utility without relying on a worst-case approximation guarantee.\footnote{Granted, since we rely on the $1$-cluster algorithm then there is a dependency between $\phi^2$ and the approximation quality of the $1$-cluster algorithm. Yet it is still possible that one would improve on the $1$-cluster algorithm's guarantees in a manner which does not end up improving upon the approximation constants of the private $k$-means approximation of Kaplan and Stemmer~\cite{KaplanSt18}.}

Before presenting the construction, we introduce the 1-cluster problem. Given a set of $n$ points in the Euclidean space $\R^d$ and an integer $t\leq n$, the goal in the 1-cluster problem is to find a ball of smallest radius $r_{opt}$ enclosing at least $t$ input points. Formally,

\begin{definition}\label{def:oneCluster}
A {\em 1-cluster problem} $(\XXX^d,n,t)$ consists of a $d$-dimensional domain $\XXX^d$ and parameters $n\geq t$.
We say that algorithm $\cal M$ solves $(\XXX^d,n,t)$ with parameters $(\Delta,w,\beta)$ if for every input database $S\in (\XXX^d)^n$ it outputs, with probability at least $1-\beta$, a center $c$ and a radius $r$ such that (i) the ball of radius $r$ around $c$ contains at least $t-\Delta$ points from $S$; and (ii) $r\leq w\cdot r_{opt}$, where $r_{opt}$ is the radius of the smallest ball in $\XXX^d$ containing at least $t$ points from $S$.
\end{definition}

Nissim and Stemmer~\cite{NS18_1Cluster} (building on~\cite{NSV16}) presented an algorithm for the 1-cluster problem with the following guarantees.

\begin{theorem}[\cite{NSV16,NS18_1Cluster}]\label{thm:1cluster}
Let $\xi>0$ be an arbitrarily small (fixed) constant.
Let $n,t,\beta,\eps,\delta$ be s.t.\ 
$$t\geq O\left(\frac{n^{\xi}\cdot\sqrt{d}}{\eps}\log\left(\frac{1}{\beta}\right)\log\left(\frac{nd}{\beta\delta}\right)\sqrt{\log\left(\frac{1}{\beta\delta}\right)}\cdot 9^{\log^*(2|\XXX|\sqrt{d})}\right).$$
There exists %a $\poly(n,d,\log(|X|/\beta))$-time 
an $(\eps,\delta)$-differentially private algorithm
that solves the 1-cluster problem $(\XXX^d,n,t)$ with parameters $(\Delta,w)$ and error probability $\beta$, where
$w=O\left( 1 \right)$
and
$$\Delta=O\left(\frac{1}{\eps}\log\left(\frac{1}{\beta\delta}\right)\log\left(\frac{1}{\beta}\right)\cdot9^{\log^*(2|\XXX|\sqrt{d})}\right).$$
%Here $n\geq t$ is the size of the input database.
\end{theorem}

In words, there exists an efficient $(\eps,\delta)$-differentially private algorithm that (ignoring logarithmic factors) is capable of identifying a ball of radius $O(r_{opt})$ containing $t-\tilde{O}(\frac{1}{\eps})$ points, provided that $t\geq\tilde{O}(n^{\xi}\cdot\sqrt{d}/\eps)$.

\begin{algorithm*}[!h]

\caption{\texttt{SampleAggregate-$k$-Means}}\label{alg:SAKmeans}

{\bf Input:} Database $X$ containing $n$ points in the $d$-dimensional ball $\BBB(0,\Lambda)$, failure probability $\beta$, and privacy parameters $\eps,\delta$.

%{\bf Tool used:} An $(\eps,\delta)$-differentially private algorithm $\AAA$ for approximating the $k$-means.

\begin{enumerate}[leftmargin=15pt,rightmargin=10pt,itemsep=1pt,topsep=0pt]

\item For $t\in[T]$ let $S_t$ be a database containing $m$ i.i.d.\ samples from $X$ (with replacement, see Theorem~\ref{thm:sampWithReplacment}), where $m=\frac{n}{2T}$ and where $T$ will be determined in the analysis.\label{alg:SAKmeans:step1}

\item For $t\in[T]$ apply a (non-private) $k$-means approximation algorithm on $S_t$ to obtain a set of $k$ centers $\check{C}_t$ such that, assuming that $S_t$ is well-separated, $\cost_{S_t}(\check{C}_t)\leq(1+\alpha)\opt_k(S_t)$. For simplicity we set $\alpha=1$.\label{alg:SAKmeans:step2}

\item Let $\check{C}=\check{C}_1\cup \check{C}_2\cup\dots\cup \check{C}_T$. 
Let $G$ be a finite grid on $\BBB(0,\Lambda)$, with grid steps $\frac{\Lambda}{nd}$. For every $t\in[T]$, let $\widetilde{C}_t$ be a set containing, for each $\check{c}\in\check{C}_t$, the closest grid point to $\check{c}$. Let $\widetilde{C}=\widetilde{C}_1\cup\dots\cup\widetilde{C}_T$.\label{alg:SAKmeans:step3}

\item Let $B=\emptyset$. For $j\in[k]$ \label{alg:SAKmeans:step4}
\begin{enumerate}
	\item Privately identify a ball of (approximately) smallest radius that encloses $\gtrsim T$ points in $\widetilde{C}$ using the algorithm from Theorem~\ref{thm:1cluster}, with privacy parameters $\left(\frac{\eps}{2k\sqrt{2k\ln(\frac{2}{\delta})}},\frac{\delta}{2k^2\cdot e^{\eps}}\right)$. Denote the center of the identified ball as $b_j$, and add $b_j$ to $B$.
	\item Delete the $T$ closest points to $b_j$ from $C$.
\end{enumerate}

\item For $i\in[k]$ let $\hat{D}_i=\min_{j\neq i}\|b_i-b_j\|$.\label{alg:SAKmeans:step5}

\item For $i\in[k]$ let $\hat{X}_i=\{x\in X : \|x-b_i\|\leq\hat{D}_i/3\}$. \label{alg:SAKmeans:step6}

\item Let $\overline{C}=\{\overline{c}_1,\dots,\overline{c}_k\}$ denote the average of the points in $\hat{X}_1,\dots,\hat{X}_k$, respectively. For $i\in[k]$ use the Gaussian mechanism with privacy parameters $(\eps,\delta)$ to compute a noisy estimation $\hat{c}_i$ of $\overline{c}_i$.\label{alg:SAKmeans:step7}

\item Output $\hat{C}=\{\hat{c}_1,\dots,\hat{c}_k\}$.\label{alg:SAKmeans:step8}

\end{enumerate}
\end{algorithm*}

We are now ready to present our variation of the sample-and-aggregate based approach of~\cite{HuangL18}. The construction is given in Algorithm \texttt{SampleAggregate-$k$-Means}. 
Consider the execution of \texttt{SampleAggregate-$k$-Means} on a database $X$ containing $n$ points in the $d$-dimensional ball $\BBB(0,\Lambda)$, and define the following good event.

\begin{center}
\noindent\fboxother{
\parbox{.9\columnwidth}{
{\bf Event ${\boldsymbol{E_1}}$ (over sampling ${\boldsymbol{S_1,\dots,S_T}}$): }\\
For every $t\in[T]$ and every set $C$ containing at most $k$ centers in $\BBB(0,\Lambda)$ we have that
$$
\left|\frac{1}{m}\cdot\cost_{S_t}(C)-\frac{1}{n}\cdot\cost_X(C)\right|\leq 5\sqrt{\frac{\Lambda^2 kd}{nm}\cost_X(C)\cdot\ln\left(\frac{2ndT}{\beta}\right)}\triangleq \Delta(C).
$$
}}
\end{center}

\begin{Claim}\label{claim:Cher}
If $\opt_k(X)\geq\frac{3\Lambda^2 n k d}{m}\cdot\ln(\frac{2ndT}{\beta})$ then Event $E_1$ occurs with probability at least $1-\beta$.
\end{Claim}

\begin{proof}
Fix a set $C$ of $k$ centers, and let $S$ be a database containing $m$ i.i.d.\ samples from $X$. Using the Chernoff bound, with probability at least $1-\beta$ we have that
\begin{equation}
\left|\frac{1}{m}\cdot\cost_S(C)-\frac{1}{n}\cdot\cost_X(C)\right|\leq\sqrt{\frac{3\Lambda^2}{nm}\cost_X(C)\cdot\ln\left(\frac{2}{\beta}\right)}.
\label{eq:Cher}
\end{equation}
We now want to apply the union bound to show that inequality~(\ref{eq:Cher}) holds for every choice of $k$ centers from $\BBB(0,\Lambda)$. To that end, consider a uniform grid $G$ on $\BBB(0,\Lambda)$ with grid steps $\frac{\Lambda}{nd}$. For a set of $k$ centers $C$ in $\BBB(0,\Lambda)$, we use $C_G$ to denote $k$ centers from the grid with the smallest distances to the centers in $C$. Observe that for every database $D$ we have
\begin{equation}
\left|\cost_D(C)-\cost_D(C_G)\right|\leq\frac{\Lambda^2}{n}.
\label{eq:grid}
\end{equation}
Now, assuming that $\opt_k(X)\geq\frac{3\Lambda^2 n k d}{m}\cdot\ln(\frac{2nd}{\beta})$, using the union bound, with probability at least $(1-\beta)$ for every choice $C_G$ of $k$ centers from the grid we have that
\begin{equation}
\left|\frac{1}{m}\cdot\cost_S(C_G)-\frac{1}{n}\cdot\cost_X(C_G)\right|\leq\sqrt{\frac{3\Lambda^2 kd}{nm}\cost_X(C_G)\cdot\ln\left(\frac{2nd}{\beta}\right)}.
\label{eq:Cher2}
\end{equation}
Combining inequalities~(\ref{eq:grid}) and~(\ref{eq:Cher2}), with probability at least $(1-\beta)$, for every choice $C$ of $k$ centers from $\BBB(0,\Lambda)$ we have
$$
\left|\frac{1}{m}\cdot\cost_S(C)-\frac{1}{n}\cdot\cost_X(C)\right|\leq5\sqrt{\frac{\Lambda^2 kd}{nm}\cost_X(C)\cdot\ln\left(\frac{2nd}{\beta}\right)}.
$$
The claim now follows from a union bound over the $T$ subsamples $S_1,\dots,S_T$.
\end{proof}

For Claim~\ref{claim:StableSubsample} and Claim~\ref{claim:SubsampleCost} we denote the following notation:
\begin{definition}[Sample-Database costs difference]\label{def:sampleRealCostsBound}
Let $n,d,k,m,T\in \N^+$, $0 \leq \beta\leq 1$, $X\in (\R^d)^n$. For any set $C\in (\R^d)^k$ define the following:
$$\Delta(C) \triangleq 5\sqrt{\frac{\Lambda^2 kd}{nm}\cost_X(C)\cdot\ln\left(\frac{2ndT}{\beta}\right)}.$$
\end{definition}
We next use Claim~\ref{claim:Cher} to show that the subsamples of a separable instance are also separable.

\begin{center}
\noindent\fboxother{
\parbox{.9\columnwidth}{
{\bf Event ${\boldsymbol{E_2}}$ (over sampling ${\boldsymbol{S_1,\dots,S_T}}$): }\\
For every $t\in[T]$ we have that $S_t$ is $2\phi$-separable.
}}
\end{center}

\begin{Claim}\label{claim:StableSubsample}
If $X$ is $\phi$-separable, and if $\opt_k(X)\geq 100\frac{n}{m}\Lambda^2 kd\cdot\ln(\frac{2nd}{\beta})$, then Event $E_2$ occurs with probability at least $1-\beta$. 
\end{Claim}

\begin{proof}
Assume that Event $E_1$ occurs, and fix $t\in[T]$. Let $C_X^{*k}$ denote an optimal set of $k$ centers for $X$. Similarly denote $C_X^{*k-1},C_{S_t}^{*k},C_{S_t}^{*k-1}$. Observe that
\begin{align*}
\frac{\opt_k({S_t})}{\opt_{k-1}({S_t})}
&=\frac{\frac{1}{m}\cost_{S_t}(C_{S_t}^{*k})}{\frac{1}{m}\cost_{S_t}(C_{S_t}^{*k-1})}\\[0.5em]
&\leq\frac{\frac{1}{m}\cost_{S_t}(C_X^{*k})}{\frac{1}{m}\cost_{S_t}(C_{S_t}^{*k-1})}\\[0.5em]
&\stackrel{(1)}{\leq}\frac{\frac{1}{n}\cost_{X}(C_X^{*k})+\Delta(C_X^{*k})}{\frac{1}{n}\cost_X(C_{S_t}^{*k-1})-\Delta(C_{S_t}^{*k-1})}\\[0.5em]
&=\frac{2\opt_k(X)+2n\Delta(C_X^{*k})}{2\cost_X(C_{S_t}^{*k-1})-2n\Delta(C_{S_t}^{*k-1})}\\[0.5em]
&\stackrel{(2)}{\leq}\frac{4\opt_k(X)}{\cost_X(C_{S_t}^{*k-1})}\\[0.5em]
&\leq\frac{4\opt_k(X)}{\cost_X(C_X^{*k-1})}\\[0.5em]
&=\frac{4\opt_k(X)}{\opt_{k-1}(X)}=4\cdot\phi^2,
\end{align*}
where Inequality (1) is due to applying Claim~\ref{claim:Cher} on the nominator and the denominator, and Inequality (2) is true when the following requirements holds: $n\Delta(C_X^{*k}) \leq \opt_k(X)$ and $2n\Delta(C_{S_t}^{*k-1})\leq \cost_X(C_{S_t}^{*k-1})$ which are true whenever 
$$\opt_k(X)\geq 100\frac{n}{m}\Lambda^2 kd\cdot\ln\left(\frac{2nd}{\beta}\right).$$
\end{proof}

%\osnote{What is $\alpha$? Why do we need $\alpha$? Why not just have $\alpha=1$?}
\begin{center}
\noindent\fboxother{
\parbox{.9\columnwidth}{
{\bf Event ${\boldsymbol{E_3}}$ (over sampling ${\boldsymbol{S_1,\dots,S_T}}$): }\\
For every $t\in[T]$ we have that $\widetilde{C}_t$ is such that $\cost_X(\widetilde{C}_t)\leq 10 \cdot \opt_k(X)$.
}}
\end{center}

\begin{Claim} \label{claim:SubsampleCost}
If $X$ is $\phi$-separable, and if $\opt_k(X)\geq\frac{100\Lambda^2 kdn}{m}\cdot\ln(\frac{2ndT}{\beta})$, then Event $E_3$ occurs with probability at least $1-\beta$. 
\end{Claim}

\begin{proof}
We first argue about $\check{C}_1,\dots,\check{C}_T$ (i.e., before the discretization). Assume that Event $E_1\wedge E_2$ occurs. Hence, each $\check{C}_t$ constructed in Step~2 satisfies $\cost_{S_t}(\check{C}_t)\leq2\cdot\opt_k(S_t)$. Fix $t\in[T]$. We have that
\begin{align*}
\cost_X(\check{C}_t)&\leq\frac{n}{m}\cdot\cost_{S_t}(\check{C}_t)+n\cdot\Delta(\check{C}_t)\\
&\leq\frac{n}{m}2\cdot\cost_{S_t}(C_{S_t}^{*k})+n\cdot\Delta(\check{C}_t)\\
&\leq\frac{n}{m}2\cdot\cost_{S_t}(C_{X}^{*k})+n\cdot\Delta(\check{C}_t)\\
&\leq\frac{n}{m}2\cdot\left[\frac{m}{n}\cost_{X}(C_{X}^{*k})+m\cdot\Delta(C_{X}^{*k})\right]+n\cdot\Delta(\check{C}_t)\\
&=2\cdot\opt_k(X)+2n\cdot\Delta(C_{X}^{*k})+n\cdot\Delta(\check{C}_t)\\
&\leq4\cdot\opt_k(X)+n\cdot\Delta(\check{C}_t),
\end{align*}
where the last inequality is because if $\opt_k(X)\geq\frac{25\Lambda^2 kdn}{m}\cdot\ln(\frac{2ndT}{\beta})$ then $n\cdot\Delta(C_{X}^{*k})\leq\opt_k(X)$. Multiplying the last inequality by 2, we get that
$$
2\cost_X(\check{C}_t)-2n\cdot\Delta(\check{C}_t)\leq 8\cdot\opt_k(X).
$$
Now if $\opt_k(X)\geq\frac{100\Lambda^2 kdn}{m}\cdot\ln(\frac{2ndT}{\beta})$ then $2n\cdot\Delta(\check{C}_t)\leq\cost_X(\check{C}_t)$, and so
$$
\cost_X(\check{C}_t)\leq 2\cost_X(\check{C}_t)-2n\cdot\Delta(\check{C}_t)\leq 8\cdot\opt_k(X).
$$
Finally,
$\cost_X(\widetilde{C}_t)\leq\cost_X(\check{C}_t)+3\Lambda^2/n\leq 10\cdot\opt_k(X)$.
\end{proof}

\begin{Claim}\label{claim:alter}
Let $\phi$ be a (sufficiently small) constant. Assume that $X$ is $\phi$-separable, and that $\opt_k(X)\geq\frac{100\Lambda^2 kdn}{m}\cdot\ln(\frac{2ndT}{\beta})$. Let $C^*=\{c^*_1,\dots,c^*_k\}$ be an optimal set of centers for $X$. For $i\in[k]$ denote $D_i=\min_{j\neq i}\|c^*_i-c^*_j\|$. With probability at least $1-\beta$, for every optimal center $c^*_i$, the set $B$ constructed in Step~4 contains a distinct center within distance $O(\gamma\cdot D_i)$ from $c^*_i$, for $\gamma=\sqrt{\frac{160\phi^2}{1-4\phi^2}}$.
\end{Claim}

\begin{proof}
Assuming that Event $E_1\wedge E_2\wedge E_3$ has occurred, and assuming that $\phi$ is small enough, by Theorem~\ref{thm:ostKmeans}, for every $i\in[k]$ there are $T$ distinct points in $\widetilde{C}$ within distance $\sqrt{\frac{160\phi^2}{1-4\phi^2}}\cdot D_i\triangleq \gamma\cdot D_i$ from $c^*_i$. We refer to these points as the {\em neighborhood} of $c^*_i$. Observe that every point in $\widetilde{C}$ is in the neighborhood of exactly one optimal center.

Now assume that, by induction, after $j$ iterations, there is a subset $J\subseteq[k]$ of size $|J|=j$ such that $B$ contains a distinct center within distance $O(\gamma\cdot D_i)$ from $c^*_i$ for each $i\in J$. In this case, assuming that $\gamma=\gamma(\phi)$ is small enough, the points that were deleted from $\widetilde{C}$ during the first $j$ iterations are exactly all of the points in the neighborhood of $c^*_i$ for every $i\in J$. Now let $\ell=\argmin_{i\notin J}\{D_i\}$, and observe that after these $j$ iterations the set $B$ contains all of the neighborhood of $c^*_{\ell}$, i.e., contains $T$ points within distance $\gamma\cdot D_{\ell}$ from $c^*_{\ell}$. Hence, by Theorem~\ref{thm:1cluster}, assuming that $T\geq\tilde{O}\left(\frac{k^{1.51}\cdot d^{0.51}}{\eps^{1.01}}\right)$, the center $b_{j+1}$ identified during iteration $(j+1)$ is such that the distance from it to its nearest point in $\widetilde{C}$ is at most $O(\gamma\cdot D_{\ell})$. Let $x$ be the closest point in $\widetilde{C}$ to $b_{j+1}$, and suppose that $x$ is in the neighborhood of $c^*_{\ell'}$ for some $\ell'\notin J$. Then, the distance from $b_{j+1}$ to $c^*_{\ell'}$ is at most $O(\gamma\cdot D_{\ell} + \gamma\cdot D_{\ell'})\leq O(\gamma\cdot D_{\ell'})$, as required.
\end{proof}

%\begin{remark}To apply Theorem~?? in the above proof, we need to bound the minimal distance between pairs of different centers in $\widetilde{C}$. To that end, we can use our assumption that the input points in $X$ come from a finite discretization of the $d$-dimensional ball $\BBB(0,\Lambda)$, with grid steps $\frac{\Lambda}{nd}$. In particular, this means that the distance between every pair of different optimal centers is at least $\frac{\Lambda}{n^3 d}$.\end{remark}

The rest of the analysis of Algorithm \texttt{SampleAggregate-$k$-Means} continues almost identically to the analysis of Algorithm \texttt{Private-Stable-$k$-Means} from Section~\ref{sec:main}. Specifically, the only property of the set $B$ that we needed in Section~\ref{sec:main} is that the distance from each optimal center to its corresponding center in $B$ is small, and this is guaranteed by Claim~\ref{claim:alter}. We obtain the following theorem.\footnote{Recall that we require $\opt_k(X)\gtrsim \Lambda^2 kd\cdot \frac{n}{m}\approx \Lambda^2 kd\cdot T\gtrsim \frac{k^{2.51}\cdot d^{1.51}\cdot \Lambda^2}{\eps^{1.01}}.$}

\begin{theorem}\label{lem:alternativeGoodC}
If $X$ is $\phi$-separable for sufficiently small $\phi$, and if $\opt_k(X)\geq$\\   $\tilde{O}_{\beta,\delta}\left(\frac{1}{\eps^{1.01}}\cdot k^{2.51}\cdot d^{1.51}\cdot \Lambda^2\right)$, then Algorithm \texttt{SampleAggregate-$k$-Means} returns w.h.p.\ a set of centers $\hat{C}$ such that
$$\cost_X(\hat{C})\leq \left(1+O(\phi^2)\right)\cdot\opt_k(X) + O(1)\cdot  \frac{k \Lambda^2 \sqrt{d}}{\eps}\cdot\ln\left(\frac{dk}{\beta\delta}\right).$$
\end{theorem}

Note that this theorem is strictly weaker than our construction from Corollary~\ref{cor:main} (because this theorem requires $\opt_k(X)$ to be much bigger than what is needed in Corollary~\ref{cor:main}). Nevertheless, we believe that this alternative construction could still be of interest for future works, as it involves different ideas than those used for obtaining Corollary~\ref{cor:main}.\\

%The analysis of the privacy properties of Algorithm \texttt{SampleAggregate-$k$-Means} is done by relating to two parts of the algorithm: first part -- step~\ref{alg:SAKmeans:step1}, and second part -- the rest of the algorithm, i.e. steps~\ref{alg:SAKmeans:step2} to~\ref{alg:SAKmeans:step8}. In the second part of the algorithm, we use simple composition theorem \ref{thm:compSimple} to sum the parameters of two serially applied mechanisms: \\

We now proceed with the privacy analysis.

\begin{theorem}
Algorithm \texttt{SampleAggregate-$k$-Means} is $\left(O(\eps),O(\delta)\right)$-differentially private.
\end{theorem}

\begin{proof}
We first argue that the set $B$ computed in Step~\ref{alg:SAKmeans:step4} is the result of a differentially private computation. To that end, let us consider an algorithm, denoted as Algorithm $\BBB$, that consists of steps~\ref{alg:SAKmeans:step2}-\ref{alg:SAKmeans:step4} of Algorithm \texttt{SampleAggregate-$k$-Means}, where the input to Algorithm $\BBB$ is $S=(S_1,\dots,S_T)$. Observe that by Theorem~\ref{thm:sampWithReplacment} in order to show that the set $B$ is the result of a differentially private computation, it suffices to show that Algorithm $\BBB$ is differentially private. Formally, steps~\ref{alg:SAKmeans:step1}-\ref{alg:SAKmeans:step4} of Algorithm \texttt{SampleAggregate-$k$-Means} can be described as sampling (with replacement) $|X|/2$ elements from the database $X$ and running $\BBB$ on the resulting sample. Hence, by Theorem~\ref{thm:sampWithReplacment}, if Algorithm $\BBB$ is differentially private, then so are the first 4 steps of Algorithm \texttt{SampleAggregate-$k$-Means}.

Now, Algorithm $\BBB$ is differentially private as it is an instantiation of the sample-and-aggregate framework of~\cite{NRS07}. In more detail, from each $S_i$ we compute (in a non-private manner) set of points $\Tilde{C}_i$, and aggregate the collection of these sets $\Tilde{C}=\bigcup_i \Tilde{C}_i$ with differential privacy. Note that a change in a single point in the database $S=(S_1,\dots, S_T)$ may lead to a change of $k$ points in the corresponding set $\Tilde{C}_i$ (and thus may affect $k$ points in $\Tilde{C}$). Recall that we aggregate the set $\Tilde{C}$ using $k$ applications of the algorithm from Theorem~\ref{thm:1cluster}. Using advanced composition (Theorem~\ref{thm:compAdvanced}) and group privacy (Theorem~\ref{thm:groupPrivacy}), these $k$ applications together satisfy $(\eps,\delta)$-differential privacy (w.r.t. $S$). This shows that Algorithm $\BBB$ is $(\eps,\delta)$-differentially private, and hence, the outcome of the first 4 steps of Algorithm \texttt{SampleAggregate-$k$-Means} is $\left(O(\eps),O(\delta)\right)$-differentially private.

The following steps of Algorithm \texttt{SampleAggregate-$k$-Means}, consist of $k$ applications of the Gaussian mechanism to disjoint parts of the input $X$. The privacy guarantees of \texttt{SampleAggregate-$k$-Means} therefore follow from simple composition.
\end{proof}

As in the analysis of Wasserstein distance for the main result, instantiating Theorem~\ref{thm:HuangL} with above {\em SampleAggregate-k-means} algorithm for the centralized model (Theorem~\ref{lem:alternativeGoodC}), we get the following result.

\begin{theorem}\label{thm:alternativeWD}
There exists an $(\eps,\delta)$-differentially private algorithm such that the following holds. 
Let $X$ be a database containing $n$ points in the $d$-dimensional ball $\BBB(0,\Lambda)$.
Assume that $X$ is $\phi$-separated for $k$-means for $\phi=O(1)$ (sufficiently small), and assume that
$$\opt_{k}(X)\geq\tilde{O}\left(
\frac{\Lambda^2}{\eps} \cdot
  \left( 
        \frac{k\sqrt{d}}{\phi^2} + 
        \frac{k^{2.51}d^{1.51}}{\eps^{0.01}}
  \right)
\right).$$
When applied to $X$, the algorithm returns, with probability at least $(1-\beta)$, a set of $k$ centers $C'$ satisfying $\wdist(C^*,C')\leq O(\phi^2\cdot\Lambda)$, where $C^*$ are the optimal centers.
\end{theorem}

%%%%%%%%%%%%%%%%%%%%%%%%%%%%%%%
% Discussion and Open Problems
%%%%%%%%%%%%%%%%%%%%%%%%%%%%%%%
\newpage
\section{Discussion and Open Problems}

This work establishes a new baseline for privately clustering stable instances, that outperforms all three existing DP-algorithms by a significant gap. 
 %improving previous results~\cite{NRS07,Wang2015,HuangL18} on multiple axes. %More importantly, our work shows how \emph{extremely} simple algorithms %(off the shelf approximation of $k$-means, sample-and-aggregate) 
%are useful in clustering privately stable instances. 
 More importantly, our work emphasizes the importance of ``simplicity'' in the design of DP clustering algorithms: Even though our algorithm is simple, and relies on folklore ideas that date all the way back to Ostrovsky et al.~\cite{DBLP:journals/jacm/OstrovskyRSS12}, it yields a significant improvement over the three existing algorithms~\cite{NRS07,Wang2015,HuangL18} in cost, in various bounds, and in portability to other problems ($k$-median) and other settings (LDP).

%We believe it is of importance not because it bring forth new algorithms or techniques, but rather
%because existing work~\cite{HuangL18} fails to provide the full details of the various guarantees obtained by existing differentially-private techniques. Furthermore, despite its title, the existing work is suboptimal on numerous axes (sample complexity bounds, additive error guarantees etc.), and unfortunately also suffers from several inaccuracies. 
%We believe that  our work helps setting the true current baseline.

Naturally, several important open problems arise from our work. %We hope that by putting forward these research directions we will assist in directing 
First, 
we pose the problem of finding a PTAS for $k$-means under stability assumptions. Non-privately, there are several papers proposing such clustering algorithms~\cite{AwasthiBS10, Cohen-AddadS17} and other works that approximate the target clustering point-wise~\cite{BalcanBG09}; whereas privately we are only able to derive a $(1+O(\phi^2))$-approximation for the $k$-means cost of $\phi$-well separated instances. In other words, in the non-private settings the quality of the approximation is independent of the input's stability guarantee, whereas in the private setting a high-quality approximation requires a very strong separation guarantee on the input. What prevents us from deriving private analogues of the above-mentioned PTASs which get a $(1+\alpha)$-approximation for any arbitrarily small $\alpha$? The reason lies in designing a private analogue to one of the most classical approaches for $k$-means approximation~--- sampling~\cite{Inaba94}. It is a well-known fact that the centroid obtained by randomly sampling $O(1/\alpha)$ datapoints from a cluster yields a $(1+\alpha)$-approximation to the cluster's cost, and the above-mentioned PTASs rely on this fact. On a high-level, a PTAS for stable inputs works by partitioning the clusters into two types: ``cheap'' clusters that cost at most $O(\alpha\phi^2\cdot \opt)$ vs ``expensive'' (non-cheap) clusters. Approximating the center of a cheap cluster relies on the notion of a core and can be made private using the $1$-cluster algorithm, but the difficulty lies in approximating the centers of the expensive clusters. In the non-private setting expensive clusters are simple to handle~--- since there are at most $O(1/\alpha\phi^2)$ such clusters, one just brute-force tries all possible centers for all expensive clusters. Alas, we have \emph{no private analogue for this approach}. More specifically, should we wish to handle expensive clusters similarly, then we first need to devise a differentially-private analogue of the PTAS of Inaba et al.~\cite{Inaba94} which runs in $n^{O(k/\alpha)}$-time. Alternatively, one could potentially derive additional properties of expensive clusters which would allow us to approximate their centers privately; or potentially try a different approach, one that doesn't rely on the separation into cheap vs.~heavy clusters.

On a related note, we also pose the question of private local-search algorithms for the clustering problem. %The local-search has been repeatedly shown to be applicable in deriving good approximation guarantees for clustering problems, both in the worst-case~\cite{Arya:2004,KanungoMNPSW04,Kanat08}, in constant dimensions~\cite{FriggstadRS16} and under stability assumptions~\cite{Cohen-AddadS17}.
 The local-search takes in addition to the input a set of candidate centers, starts with an arbitrary $k$-size subset of centers and then repeatedly replaces one (or a few) of the centers with other candidate centers if they improve the %$k$-means / facility-location
 cost significantly. While the step of center replacing can be done privately, it requires that the set of candidate centers %/ facilities 
be public~\cite{GuptaLMRT10}. Of course, in the non-private setting one may use the datapoints themselves as the set of candidate centers;
%(or all centroids obtained from all subsets of size $O(1/\alpha)$ of datapoints);
yet, despite the works of~\cite{DBLP:conf/icml/BalcanDLMZ17, KaplanSt18}, we do not know how to get a {\em small} set of candidate centers with differential privacy (that contains a subset of $k$ centers whose cost is no more than $(1+\alpha)$ times the optimal cost). 
%To derive a set of candidate centers privately is an important open problem. Note that existing works on private coresets~\cite{FFHN09} are not guaranteed to produce good centers, rather just answer any tuple of $k$-centers with fairly good accuracy.

Third, 
%it is quite possible that, 
despite the fact that the $k$-means and $k$-median problems are part of the ``CS-canon'', it is possible these two problems are the ``wrong'' problems to approximate with a differentially private algorithm. The reason lies in the sensitivity of the optimal $k$ centers, even for stable instances. %Huang and Liu presented a $\phi$-well separated instance where the optimal $k$-centers shift by $O(\phi^2\Lambda)$ 
%($\Lambda$ denoting the diameter of the instance) 
%in Wasserstein distance with the addition of one datapoint. Such a shift is due to multiple points that are almost as close to $2$ (or more) centers.
 However, if instead of outputting the ``true'' $k$-means centers we shift our focus to outputting some notion of ``core centers'' or centers that best represent the fraction of the instance with clear preference among centers\footnote{Note how this proposed ``definition'' is recursive and thus ill-defined.}, then such objectives might be less sensitive to a change of a single datapoint and could therefore be better suited for differential privacy. In essence, we re-pose the question of a definition of clustering which is generalizable. 

%Last, we pose the question of better private approximation for the $1$-median problem in the Euclidean space. In our work we ``plugged-in'' the guarantees of the private gradient descent algorithm~\cite{BST14}, but it is likely this approach gives a suboptimal utility guarantee. We do not know of any work focusing on the private geometric median problem and we believe it is an interesting direction to explore, in particular in terms of lower bounds. One promising venue for such an algorithm could be the simple algorithm in~\cite{CohenLMPS16}.

%%%%%%%%%%%%%%%
% bibliography
%%%%%%%%%%%%%%%
\bibliographystyle{unsrt} 
\bibliography{M_Sc_thesis}

%%%%%%%%%%%%%%%
% Appendix
%%%%%%%%%%%%%%%
\appendix

\section{Proof of Theorem~\ref{thm:ostMedian} (\cite[Theorem 5.1]{DBLP:journals/jacm/OstrovskyRSS12} for $k$-median)}
\label{apx_sec:ostrovskythmproof}

For completeness, we include here the proof of~\cite{DBLP:journals/jacm/OstrovskyRSS12} for Theorem~\ref{thm:ostMedian} (the $k$-median version of Theorem~\ref{thm:ostKmeans}). We first restate the theorem.

\begin{theorem}[{{\cite[Theorem 5.1]{DBLP:journals/jacm/OstrovskyRSS12} for $k$-median}}]\label{apx_thm:ostMedian}
	Let $\alpha$ and $\phi$ be such that $\frac{\alpha+\phi}{1-\phi}<\frac{1}{4}$. Suppose that $X\subseteq\R^d$ is $\phi$-separated for $k$-median, let $C=(c_1,\dots,c_k)$ be a set of optimal centers for $X$, and let $\hat{C}=(\hat{c}_1\dots,\hat{c}_k)$ be centers such that $\medcost_X(\hat{C})\leq\alpha\cdot\medopt_{k-1}(X)$. Then for each $\hat{c}_i$ there is a distinct optimal center, call it $c_i$, such that $\|\hat{c}_i-c_i\|\leq2\cdot\frac{\alpha+\phi}{1-\phi}\cdot D_i$, where $D_i = \min_{j\neq i}\|c_i-c_j\|$.
\end{theorem}

\begin{proof}
	Let $\rho=(\frac{\alpha}{\phi}+1)^{-1}$. For $i\in[k]$ define $r_i={\frac{1}{n_i}\sum_{x\in X_i}\|x-c_i\|}$, and $X_i^{\rm cor}=\{x\in X_i: \|x-c_i\|\leq\frac{r_i}{{\rho}}\}$. 
	A standard argument (see Claim~\ref{clm:markovArg}) shows that $|X_i^{\rm cor}|\geq (1-\rho)n_i$. Let $d_i=\phi\opt_{k-1}(X)/n_i$. We argue that $r_i\leq d_i\leq\frac{\phi}{1-\phi}\cdot D_i$. Indeed,
	$$
	r_i=\frac{1}{n_i}\sum_{x\in X_i}\|x-c_i\|\leq\frac{1}{n_i}\cdot\opt_k(X)\leq\frac{1}{n_i}\phi\opt_{k-1}(X)=d_i,
	$$
	which shows the first inequality. To see that $d_i\leq\frac{\phi}{1-\phi}\cdot D_i$, we first show the following:
	$$
	\opt_{k-1}(X)\leq\opt_k(X)+n_i\cdot D_i
	$$
Above holds since by triangle inequality the right term is upper bounding a cost of the following $k-1$  centers: $C\setminus\{c_i\}$, with the following assignment: each cluster $j\in [k]\setminus\{i\}$ is assigned to its center, and the points of cluster $i$ are assigned to the closest center of $c_i$. That  assignment is a $k-1$ cost, that is lower bounded by the optimal $k-1$ cost, the right term, thus establishing the inequality.\\

Now, by the input assumption we have $\opt_k(X)+n_i\cdot D_i \leq \phi\opt_{k-1}(X)+n_i\cdot D_i$.
These two inequalities yields $\opt_{k-1}(X)\leq n_i\cdot D_i/(1-\phi)$, and hence, 
	$$
	d_i=\phi\opt_{k-1}(X)/n_i\leq \frac{\phi}{1-\phi}\cdot D_i.
	$$
	
	We say that a center $\hat{c}_i$ is {\em close} to an optimal center $c_j$ if $\|\hat{c}_i-c_j\|\leq2\cdot{\frac{\alpha+\phi}{1-\phi}}\cdot D_j$. Observe that if $2\cdot{\frac{\alpha+\phi}{1-\phi}}<1/2$ then a center $\hat{c}_i$ can be close to at most one optimal center. 
	Assume towards contradiction that there is a center $\hat{c}_i$ such that $\hat{c}_i$ is not close to any optimal center. Therefore, by the pigeonhole principle, there must exist an optimal center $c_j$ that is not close to any center in $\hat{C}$.
	Then, in the clustering around $\hat{c}_1,\dots,\hat{c}_k$, all the points in $X_j^{\rm cor}$ are assigned to a center that is more than $2\cdot{\frac{\alpha+\phi}{1-\phi}}\cdot D_j$ away from $c_j$. Recall that $X_j^{\rm cor}$ contains all points whose distance to $c_j$ is at most 
	$$\frac{r_j}{{\rho}}\leq \frac{{\frac{\phi}{1-\phi}\cdot D_j}}{{\rho}}={\frac{\alpha+\phi}{1-\phi}}\cdot D_j.$$
	Hence, in the clustering around $\hat{c}_1,\dots,\hat{c}_k$, all the points in $X_j^{\rm cor}$ are assigned to a center that is more than ${\frac{\alpha+\phi}{1-\phi}}\cdot D_j$ away from them. Therefore,
	\begin{align*}
	\cost_X(\hat{C})&> |X_j^{\rm cor}|\cdot \frac{\alpha+\phi}{1-\phi}\cdot D_j\\
	&\geq \left(1-{\rho}\right)n_j \cdot \frac{\alpha+\phi}{1-\phi}\cdot D_j\\
	&= \alpha \cdot n_j \cdot \frac{D_j}{1-\phi}\\
	&\geq \alpha \cdot n_j \cdot \frac{d_j}{\phi}\\
	&= \alpha\cdot\opt_{k-1}(X),
	\end{align*}
	giving a contradiction.
\end{proof}

\end{document}